\documentclass{vgtc}                          % final (conference style)
\ifpdf%                                % if we use pdflatex
  \pdfoutput=1\relax                   % create PDFs from pdfLaTeX
  \usepackage{graphicx}                % allow us to embed graphics files
  \DeclareGraphicsExtensions{.pdf,.png,.jpg,.jpeg} % for pdflatex we expect .pdf, .png, or .jpg files
\else%                                 % else we use pure latex
  \usepackage{graphicx}                % allow us to embed graphics files
  \DeclareGraphicsExtensions{.eps}     % for pure latex we expect eps files
\fi%

\graphicspath{{figures/}{pictures/}{images/}{./}} % where to search for the images
\usepackage{enumitem}% http://ctan.org/pkg/enumitem

\usepackage{microtype}                 % use micro-typography (slightly more compact, better to read)
\PassOptionsToPackage{warn}{textcomp}  % to address font issues with \textrightarrow
\usepackage{textcomp}                  % use better special symbols
\usepackage{mathptmx}                  % use matching math font
\usepackage{amsmath}
\usepackage{amssymb}
\usepackage{times}                     % we use Times as the main font
\usepackage{cite}                      % needed to automatically sort the references
\usepackage{tabu}                      % only used for the table example
\usepackage{booktabs}   
\usepackage{comment}
\usepackage{soul}
\usepackage{float}
\usepackage{multirow}
\usepackage{array}
\newcolumntype{L}{>{\centering\arraybackslash}m{3cm}}

\usepackage{fancyhdr} 
\usepackage[ruled,linesnumbered]{algorithm2e}
\usepackage{multicol}
\usepackage{graphicx}
\usepackage{subfig}
\usepackage[T1]{fontenc}
\usepackage{makecell}
\usepackage[normalem]{ulem}
\usepackage{xcolor}
\usepackage[titletoc,title]{appendix}

\renewcommand\theadfont{\small\bfseries}
\usepackage{amsmath}

\usepackage[colorlinks,urlcolor=blue,linkcolor=blue,citecolor=blue]{hyperref}

\usepackage{enumitem}% http://ctan.org/pkg/enumitem

\usepackage{microtype}                 % use micro-typography (slightly more compact, better to read)
\PassOptionsToPackage{warn}{textcomp}  % to address font issues with \textrightarrow
\usepackage{textcomp}                  % use better special symbols
\usepackage{mathptmx}                  % use matching math font
\usepackage{amsmath}
\usepackage{amssymb}
\usepackage{times}                     % we use Times as the main font
\usepackage{cite}                      % needed to automatically sort the references
\usepackage{tabu}                      % only used for the table example
\usepackage{booktabs}   
\usepackage{comment}
\usepackage{soul}
\usepackage{float}
\usepackage{multirow}
\usepackage{array}
\usepackage{wrapfig}
\usepackage[export]{adjustbox}
\usepackage{hyperref}
\hypersetup{
    colorlinks=true,
}

\usepackage{fancyhdr} 
\usepackage[ruled,linesnumbered]{algorithm2e}
\usepackage{multicol}
\usepackage{graphicx}
\usepackage[T1]{fontenc}
\usepackage{makecell}
\usepackage[normalem]{ulem}
\usepackage{xcolor}

\renewcommand\theadfont{\small\bfseries}

\newcommand{\yifannn}[1]{}
\usepackage{amsmath}
\newcommand{\xiaoqi}[1]{{\color{black}#1}}

\onlineid{0}

\vgtccategory{Research}

\title{DeepGD: A Deep Learning Framework for Graph Drawing Using GNN}

\author{Xiaoqi Wang\thanks{e-mail: wang.5502@osu.edu}\\ %
        \scriptsize The Ohio State University %
\and Kevin Yen\thanks{e-mail: kevinyen@verizonmedia.com}\\ %
     \scriptsize Yahoo Research%
\and Yifan Hu\thanks{e-mail: yifanh@gmail.com}\\ %
     \scriptsize Yahoo Research
\and Han-Wei Shen\thanks{e-mail: shen.94@osu.edu}\\
    \scriptsize The Ohio State University}
\abstract{In the past decades, many graph drawing techniques have been proposed for generating aesthetically pleasing graph layouts. However, it remains a challenging task since different layout methods tend to highlight different characteristics of the graphs. Recently, studies on deep learning based graph drawing algorithm have emerged but they are often not generalizable to arbitrary graphs without re-training. In this paper, we propose a Convolutional Graph Neural Network based deep learning framework, DeepGD, which can draw arbitrary graphs once trained. It attempts to generate layouts by compromising among multiple pre-specified aesthetics considering a good graph layout usually complies with multiple aesthetics simultaneously. In order to balance the trade-off, we propose two adaptive training strategies which adjust the weight factor of each aesthetic dynamically during training. The quantitative and qualitative assessment of DeepGD demonstrates that it is capable of drawing arbitrary graphs effectively, while being flexible at accommodating different aesthetic criteria.
} % end of abstract

\CCScatlist{
  Graph Drawing, Deep Learning, Convolutional Graph Neural Network, Aesthetic Criteria of Graph
}
    
\nocopyrightspace

\begin{document}

\renewcommand{\sectionautorefname}{Section}
\renewcommand{\subsectionautorefname}{Section}
\renewcommand{\subsubsectionautorefname}{Section}

\renewcommand\floatpagefraction{1}
\makeatletter
\newcommand\footnoteref[1]{\protected@xdef\@thefnmark{\ref{#1}}\@footnotemark}
\makeatother
% trim space
% \setlength\intextsep{5pt}
% \setlength\belowdisplayskip{1pt} 
% \setlength\belowdisplayshortskip{1pt}
% \setlength\abovedisplayskip{1pt} 
% \setlength\abovedisplayshortskip{1pt}


\firstsection{Introduction}

\maketitle

A graph is a mathematical structure in which nodes represent entities and edges indicate the relationships among the entities. Graphs are widely used to represent many different types of data such as transactions, transportation networks, %molecules structures,
social relationship, etc. 
%Mathematically, a graph can be formulated as an adjacency matrix whose elements indicate pair-wise relationship between the nodes. However, it is difficult to discover insights from the data by observing the adjacency matrix alone. 
The most popular way to visualize graphs is to use node-link diagrams, where the topological structure of the graph can be directly visualized. 

% In the past decades, many graph drawing techniques have been proposed to draw graphs as node-link diagrams.  Some layout methods emphasize natural groupings or semantic qualities within or between the communities \cite{CiSE}, while other methods focus on revealing the hierarchical structures \cite{treeGraph}. Also, if the chosen layout method has some hyper-parameters to tune, adopting a trail-and-error process is inevitable but time-consuming. Thus, the primary concern of research is how to find a universal algorithm to generate "good" layouts.

However, graph drawing is a challenging task. Different layout methods tend to highlight different characteristics of a graph. Therefore, choosing the most appropriate layout method usually requires in-depth knowledge about layout methods as well as the graph being drawn. This leads to a question: what's the most appropriate layout method and how should we evaluate it?
%and how to make different layout methods comparable to each other.

The goodness of a graph layout can be evaluated by 
%some 
aesthetic metrics such as edge length variation, stress, minimum angle, etc. Different metrics focus on different visual properties and cater for different human preferences. Nevertheless, there is no single consensus on which metric is the best, and some of the metrics are even contradictory to each other. Therefore, a method that considers multiple aesthetic metrics simultaneously and makes sensible compromises is more likely to generate a visually pleasing graph layout~\cite{huang-2013}. Furthermore, we are ultimately interested in learning human preference automatically. Towards that goal, we need a general framework that is able to optimize an arbitrary objective function that represents the human preference.

In recent years, new techniques have been proposed to utilize machine learning algorithms for graph layouts. However, %as a data-driven approach, machine learning's capability is constrained by data resources. Hence, 
some of these algorithms require special training data for each graph drawing task so that the model needs to be re-trained with new training data. For example, if it is trained to draw a star graph, the model cannot properly draw a tree graph without re-training on new data. To relax this constraint, %some 
other layout methods involve human interaction to collect data and to learn the graph drawing per human preference. %~\cite{bach,barosa,sebag,masui,sponemann}. Correspondingly, they require intensive labors. % to train this interactive system.
In short, among the machine learning based graph drawing algorithms proposed so far, it is not possible to train a general graph drawing model that directly optimizes multiple aesthetic criteria, as specified by a cost function.

In this paper, we propose {\em DeepGD}, a deep learning framework for graph drawing, which generates graph layouts complying with multiple aesthetic metrics simultaneously. DeepGD can be applied easily to optimize most of the commonly agreed aesthetic metrics. Also, DeepGD only needs to be trained once and can subsequently be applied to draw arbitrary types of graphs. In terms of the methodology, a Convolutional Graph Neural Network (ConvGNN) is used to produce the position of nodes such that the desired aesthetic metrics are optimized in the resulting layout. To accomplish this goal, a multi-objective loss function is designed in a way that each of the aesthetic aspects is represented by an individual component in the composite loss function. Additionally, two adaptive training strategies are proposed to automatically adjust the weight factors corresponding to each loss component such that the human preference on aesthetics trade-off is reflected in the weight factors.

The effectiveness and efficiency of the proposed deep learning framework is evaluated both qualitatively and quantitatively against baseline including the stress majorization algorithm~\cite{gansner_stress_major} and PivotMDS~\cite{PMDS}. The results from our extensive experiments show that our work can generate aesthetically pleasing graphs layout by compromising between multiple aesthetic metrics. In addition, the robust performance of our model shows that it has the ability to capture the latent patterns and relationships in graph data, instead of merely memorizing the training samples. 

The primary contributions of this work are:
\begin{itemize}[noitemsep,topsep=0pt]
  \item A novel ConvGNN-based framework for graph drawing capable of 
  incorporating multiple aesthetic metrics simultaneously.
  %and generates graph layout for any unseen graphs, without the need of re-training on a per graph basis.
%   \item A deep learning framework that can be easily reformulated to optimize most of commonly agreed aesthetic metrics.
  \item A flexible model design which makes DeepGD applicable to arbitrary types of graphs once the model is trained.
  \item Two adaptive training strategies for graph drawing that adjust the weight factors dynamically to balance the trade-off among aesthetics.
  \item A comprehensive experimental study that optimizes minimum angle, edge length variation, stress energy, t-SNE, and node occlusion loss functions, either on their own or in combination. 
  %with quantitative and qualitative 
  %compared against a traditional graph drawing algorithm.  
\end{itemize}

It is worth mentioning that scaling up for large graphs is not the focus of this paper. Rather, our work confirms that our proposed deep learning framework can produce aesthetically pleasing layouts that optimize {\em arbitrary objective functions} for small {\em general graphs}. Further discussion is given in the discussion section.

\section{Related Work}
Our work is related to three fields: graph visualization, graph neural networks, and machine learning approaches for graph visualization. The previous work in each field is discussed and summarized in this section.
\subsection{Graph Visualization}
\label{section:liter-graphvis}
%Since Tutte \cite{tutte_1963} introduced one of the first graph drawing algorithm based on barycentric representation, 
There have been a multitude of graph visualization techniques proposed in the past five decades. 
\yifannn{Gibson et al.~\cite{gibson} classify the existing techniques into three categories: dimensionality reduction based approaches, computational improvements with multi-level graph layouts, and force-directed graph layouts. 
The first category, dimensionality reduction based approaches (e.g.,

\cite{brandes_pich_2006,kruiger_2017}),
%\cite{bonabeau_2002,brandes_pich_2006,harel_koren_2004,kruiger_2017,bonabeau_1998}, 
captures information in high-dimensional space such as graph-theoretical distance between nodes, and preserve that specific information in a two-dimensional layout. The second category is computational improvements with multi-level techniques (e.g., \cite{gajer_2004,hu}). 

%harel_koren_2002,walshaw_2003, hachul_2005,
%which focuses on optimizing the graph layout in a coarser representation and then map that optimal layout back to the graph layout in the original representation. 
%Lastly, a detailed explanation and discussion for the force-directed graph layout will be given below, since its concept plays an essential role in our case study.}
}
The force-directed graph layout models the graph as a physical system in which adjacent nodes are pulled by the attractive force and all other nodes are pushed away by the repulsive force %~\cite{kobourov_2013}. There are two types of force-directed approaches: spring-embedder based approaches %\cite{eades_1984,fruchterman_reingold_1991} 
 following energy minimization principles (e.g., \cite{gansner_stress_major}). %kamada_kawai_1989,noack_2007, ForceAtlas2
%Specifically, the spring-electrical based approaches attempt to find an equilibrium in a physical system such that the total force on each node is zero, while the energy-based approaches formulate graph layout as an optimization problem which minimizes the energy function encoding the desired properties of the graph. 
\yifannn{One of the energy-based approaches is the stress model, which mimics the energy of spring forces,
and can be solved by a majorization-based optimization technique~\cite{gansner_stress_major}. Since both approaches minimize a loss function in the end, they both can fit into DeepGD, which is able to optimize any loss function.
}

%even the spring-electrical  Our deep learning frame work works with both
%In our work, we also formulate our deep learning framework to minimize the stress and achieve a comparably low stress.

To evaluate the quality of a graph layout, there are some commonly agreed aesthetic metrics, and each of which highlights different visual properties.\yifannn{\cite{huang-2015}.}
However, aesthetics metrics often conflict with each other~\cite{Nascimento}. As a result, most graph drawing algorithms aim to satisfy only one or two aesthetic criteria. It is widely recognized that a balance of aesthetics yields the best graph layout~\cite{didimo}. Hence, 
\yifannn{Argyriou et al. \cite{Argyriou} and} 
Huang et al. \cite{huang-2013} propose a force-directed approach to generate layout with the best compromise between the spring force, crossing angle force, and incident angle force. However, this force-directed approach lacks the flexibility of accommodating arbitrary combination of aesthetics. In this paper, we propose a general deep learning framework for graph drawing that is highly flexible over the choice of the target aesthetics and is capable of improving them simultaneously with adaptively adjusted weights.

\subsection{Graph Neural Network}
\label{section:gnn}

Graph Neural Network (GNN) is designed to adapt deep learning to 
the combinatorial nature of graphs. Convolutional Graph Neural Network (ConvGNNs), as one type of GNN, is employed in our work.
Encouraged by the success of Convolutional Neural Networks (CNN) in image data, the convolution operation is extended to graph data in Convolutional Graph Neural Networks (ConvGNN). In general, ConvGNNs are designed to generate node embeddings by aggregating information from their neighboring nodes. 
% Since one graph convolutional layer aggregates information from the immediate neighbors of a node, stacking up multiple graph convolutional layers can thus gather the information from non-immediate neighbors and generate node embeddings which emphasize more on the global graph structure.

ConvGNNs can be divided into two main streams: spectral-based approaches %\cite{bruna-zaremba,kipf-max,cayleynet,li-wang-zhu,zhuang-ma} 
and spatial-based approaches. %\cite{micheli-alessio,hamilton-ying,monti-boscaini,chen-ma-cao,xu-hu-jegelka}. 
%The main difference between these two main streams is how they define convolutions. Spectral-based approaches define graph convolutions as removing noises from graph signal, while spatial-based approaches define it as information propagation. 
In this work, a spatial-based convolutional layer \cite{gilmer-riley} is employed as the building block of our model architecture. The graph convolution layer aggregates neighbors' information for each node while considering the characteristics of the connection between the node and its neighbors. In our case, each message passed from a node's neighbor is weighted according to the direction and magnitude of the difference between two node embeddings.

\subsection{Machine Learning Approaches for Graph Visualization}
% The history of machine learning on graph drawing dates back to 1993. Between 1993 and 2005, some researches had been conducted on the application of machine learning for graph drawing but the capability of these machine learning approaches are limited by the available computational resources and the level of machine learning theories at that time. After 2005, no more publications related to machine learning on graph drawing can be found until 2012 \cite{vieira}. Since 2012, this field again arouse the attention of researchers along with the rapid advancement of machine learning.
% The history of machine learning on graph drawing dates back to 1993~\cite{vieira}. \yifannn{However, only a few researches in this field have been conducted} 
The applications of machine learning for graph drawing problem can be classified into three categories: graph drawing with human interaction, graph drawing without human interaction, and the evaluation of graph drawing.  

The earliest work \cite{neto_eades_1993} in the first category proposed an interactive system which automatically adjusts the objective function and the parameters of a simulated annealing graph drawing method obtained from the user preference. Since then, many other approaches aimed at adjusting the fitness function of genetic algorithms by collecting information from human feedback. Specifically, Spönemann et al. \cite{sponemann} collects human feedback by designing a slider for user to indicate the desired aesthetic criteria and a canvas to select the better layout from a collection of layouts.

% The second category is composed of various graph drawing approaches without human interaction. 
%Since these approaches fail to keep human in the loop, the subjective human preference is not considered in the generated graph layout. 
The approaches %~\cite{kwon-ma-2018,kwon-ma-2020,bernd,wang-okazaki,olver-benno,wang-qu-2019} 
in second category focus on generating graph layouts based upon the graph structure per se or the result from other traditional graph drawing methods. \xiaoqi{For example, Wang et al.~\cite{wang-qu-2019} proposed an LSTM-based neural network which learns the layout characteristics from the results of other graph drawing techniques, and draws a graph in a similar drawing style as the specific targeting technique. The limitation is that this approach requires collecting new training data and model re-training for different targeting techniques and for
different types of graphs.} Kwon and Ma~\cite{kwon-ma-2020} designed a deep generative model, which systematically draws a specific graph in diverse layouts. For each graph, a two-dimensional latent layout space is constructed that allows users to navigate and explore various layouts. Nevertheless, a model trained on one graph is only applicable to generating layouts for that specific graph. DeepGD also falls into this category. \xiaoqi{{\em However, our approach is more flexible because once the desired aesthetic criteria are specified and the model is trained, it can be applied to arbitrary types of graphs.}} Furthermore, even though the model needs to be re-trained for different aesthetic criteria, the same training data set can be reused regardless of the chosen criteria.

The last category is evaluation of graph layouts with machine learning approaches. %There are only a few studies found in this category. 
Klammler et al. \cite{mchedlidze} uses a Siamese neural network to identify a more aesthetically pleasing layout from a pair of layouts. A different evaluation approach is proposed by Haleem et al.\cite{haleem-huamin} who designs a CNN-based model to predict various aesthetic metrics for a graph layout without knowing the node and edge coordinates.

\section{Background}
This background section introduces some preliminary knowledge and basic concepts about graph drawing and ConvGNN. We represent a graph as $G = (V, E)$, where $V$ and $E$ is the set of vertices and edges. The layout is represented as  $\mathbf{x}: V\rightarrow \mathbb{R}^2$. The graph theoretic distance between nodes $u$ and $v$ is denoted as $d_{uv}$.

\subsection{Graph Drawing}
We describe two popular graph drawing algorithms including stress majorization\cite{gansner_stress_major} and t-SNE\cite{kruiger_2017} in this section. Both algorithms serves as the bases to define our loss function.

%Since the paper's main goal is to generate aesthetically pleasing graph layouts, the fundamental concepts of graph drawing pave the way for our work. Hence, this background section for graph drawing 
% We introduce stress majorization, t-SNE and several commonly agreed aesthetics for designing graph layouts. These also serve as the bases to define %Besides, all subsections are closely related to 
% our loss functions later.
\subsubsection{Stress Majorization}
\label{section:stress-major}
Stress majorization formulates a graph as a physical system in which there exists a spring between each pair of nodes. %Stress, as the potential energy stored in the spring system, is minimized so that the aesthetic quality will be improved significantly \cite{kobourov_2013}. 
The stress energy for $G$ is computed as
\vspace{0.2cm}
\begin{equation}
\centering
\label{stress}
L_\text{stress}=
\sum_{u, v \in V, u \neq v}
    w_{uv} \left(\|\mathbf{x}_u - \mathbf{x}_v\| - d_{uv}\right)^2 ,
\end{equation}
where the weighting factor $w_{uv}$ is typically  set  to $1/d_{uv}^2$.
%$\frac{1}{d_{uv}^2}$. 
%The stress becomes zero if the graph-theoretic distance $d_{uv}$ equals to the Euclidean distance $\|x_u - x_v\|$. If $d_{uv} > \|x_u - x_v\|$, a repulsive force will enforce node $u$ and $v$ to disperse more; If $d_{uv} < \|x_u - x_v\|$, an attractive force will enforce node $u$ and $v$ to get closer.

Gasner et al. \cite{gansner_stress_major} propose a stress majorization algorithm which minimizes the stress by a majorization-based optimization approach. 
%For each graph G, this approach will iteratively update every node's position until convergence. Besides, they have shown that the stress is guaranteed to strictly decrease for each iteration. 
This approach iteratively updates position of nodes as follows.
\begin{align}
\centering
\label{stress-iter}
\mathbf{p}_u \leftarrow 
    \sum_{v \neq u} w_{uv} \bigg(\mathbf{x}_v+d_{uv}\frac{\mathbf{x}_u-\mathbf{x}_v}{\|\mathbf{x}_u-\mathbf{x}_v\|}\bigg) \bigg/ \sum_{v \neq u} w_{uv}.
\end{align}

% \begin{align}
%   \frac{\partial}{\partial s} 
%   \sum_{u,v \in V, u \neq v} \frac{|sl_{uv} - d_{uv}|^2}{d_{uv}^2} = 0 \\
%     s = \frac{
%         \sum l_{uv} / d_{uv}
%     }{
%         \sum l_{uv}^2 / d_{uv}^2
%     } 
% \end{align}
\subsubsection{tsNET}
\label{sec:tsne-background}
The t-distributed stochastic neighbor embedding (t-SNE) is a dimensionality reduction algorithm widely used for visualizing various types of data. Kruiger et al.\cite{kruiger_2017} adapted t-SNE into the context of graph visualization and proposed a dimensionality reduction based graph drawing approach called tsNET. 

% It formulates graph drawing as an optimization problem such that the divergence between graph space and layout space is minimized.

tsNET aims at minimizing the divergence between the graph space and the layout space. Let $d_{ij} \in \mathbb{N}$ denotes the graph theoretic distance between node $i$ and $j$ in a graph of size $N$. Then, the graph space similarity $p_{ij}$ between node $i$ and $j$ is computed from a normalized Gaussian distribution as follows,

% Given a graph with $N$ nodes, $D \in \mathbb{R}^{N \times N}$ denotes the graph theoretic distance matrix. This distance matrix in graph space is then transformed by a normalized Gaussian distribution to obtain a probability matrix in which the conditional probability of node $i$ has node $j$ as its neighbor is computed as follows.
\begin{align}
p_{ij} = p_{ji} &= \frac{p_{j|i}+p_{i|j}}{2N} \\ 
\text{where } p_{j|i} &= \frac{\exp(-\frac{d_{ij}^2}{2\sigma_i^2})}{\sum_{\substack{k\\k\neq i}}\exp(-\frac{d_{ik}^2}{2\sigma_i^2})}
\end{align}
and $\sigma_i$ is a hyper parameter representing the Gaussian standard deviation for node $i$. 
% Then, the probability of node $i$ and $j$ becomes neighbor in the graph is simply a symmetrized version of the conditional probability $p_{j|i}$ and $p_{i|j}$, namely $p_{ij} = p_{ji} = \frac{p_{j|i}+p_{i|j}}{2N}$ ($p_{ii} = 1$).

% In the layout space, the layout of a graph G is encoded as an Euclidean distance matrix which contains the pairwise Euclidean distances between every pair of nodes. Similarly, this distance matrix is transformed by a normalized Student's t-distribution. The probability of node $i$ and $j$ stay close in layout space is computed as
Similarly, in the layout space, the similarity $q_{ij}$ between node $i$ and $j$ is derived from a t-distribution as follows,
\begin{equation}
    q_{ij} = q_{ji} = \frac{(1+||\mathbf{x}_i-\mathbf{x}_j||^2)^{-1} }{\sum_{\substack{k,l\\k \neq l}}(1+||\mathbf{x}_k-\mathbf{x}_l||^2)^{-1}}.
\end{equation}

As a result, the optimal layout can be obtained by minimizing the KL-divergence between $p_{ij}$ and $q_{ij}$, namely
\vspace{0.2cm}
\begin{align}
    L_\text{t-SNE} = \sum_{\substack{i,j\\i \neq j}} p_{ij}\log\frac{p_{ij}}{q_{ij}}.
    \label{equ:tsne-loss}
\end{align}

\subsection{Convolutional Graph Neural Network}
\label{section:background-gnn}
The proposed deep learning framework is built upon a Convolutional Graph Neural Network (ConvGNN) which utilizes the convolution operation to aggregate messages from neighboring nodes. Specifically, the spatial-based convolutional network \cite{gilmer-riley} lays the foundation of our model architecture, and is described as follows. 

Given a graph with $N$ node and $D$ node features, a ConvGNN takes a node feature matrix $\widetilde{\mathbf{X}} \in \mathbb{R}^{N \times D}$ and an adjacency matrix $\mathbf{A} \in \mathbb{R}^{N \times N}$ as inputs. A hidden layer $i$ with output feature length $F^i$ in the ConvGNN can thus be written as,
\begin{equation}\label{gnn-prop}
    \begin{cases}
        \mathbf{H}^0& =\   \widetilde{\mathbf{X}},\\
        \mathbf{H}^i& = \ f(\mathbf{H}^{i-1}, \mathbf{A}),
    \end{cases}
\end{equation}
where $\mathbf{H}^i \in \mathbb{R}^{N \times F^i}$ is the hidden node representation after $i^{th}$ layer. The function $f$ is the propagation rule of choice, also known as the the message aggregation function, which determines how the messages are passed between connected nodes.

\begin{figure}
  \begin{center}
    \includegraphics[width=\linewidth]{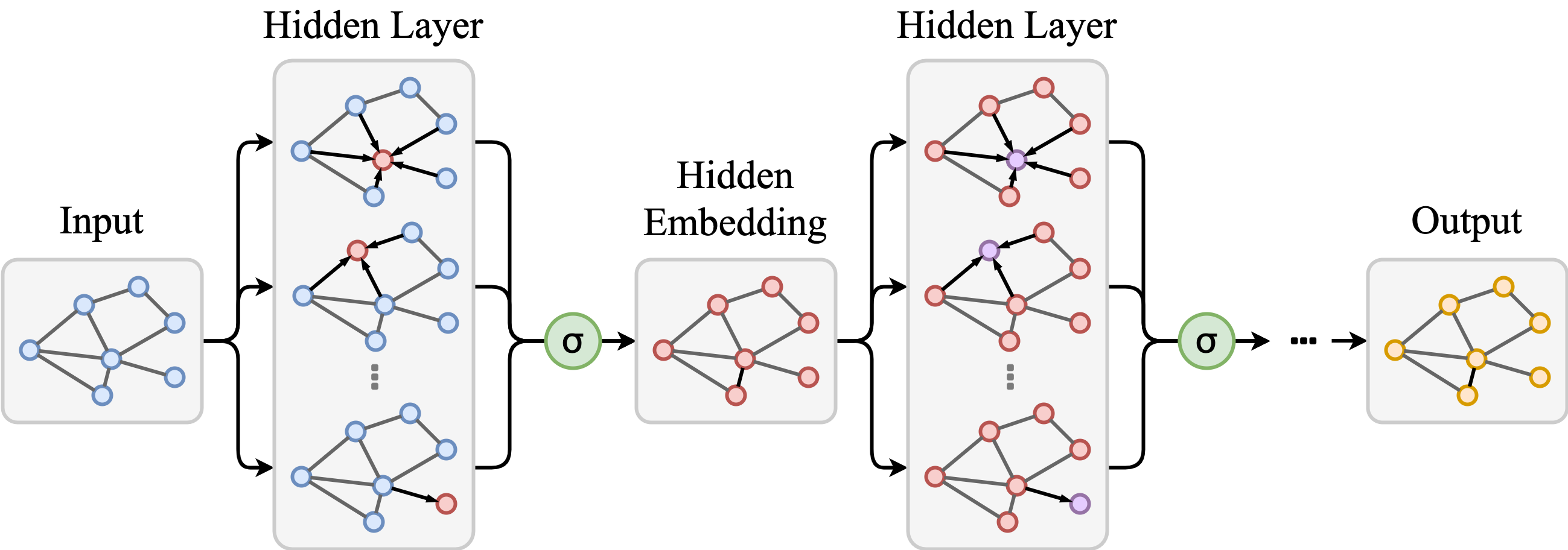}
  \end{center}
  \caption{A multi-layer ConvGNN. In each hidden layer, arrow represents the direction of message flow and the change of node color indicates that the hidden node representation is updated by the aggregator function.}
  \label{gnn-explain}
\end{figure}

In general, as shown in \autoref{gnn-explain}, all propagation rules follow a basic idea -- the hidden representation of each node in the current layer is aggregated from the hidden representation of its neighbors in the previous layer. In this case, after $i$ layers, each node can gather information from nodes that are $i$ graph-theoretic distance away. At the end of training, ConvGNN learns a set of aggregator functions that define the way of combining feature information from a node's local neighborhood \cite{hamilton-ying}. During the inference phase, the network utilizes learned aggregators to generate meaningful embeddings for unseen graphs.

\section{Methodology}
\label{section:approach}
We propose a general deep learning framework for generating graph layouts complying with multiple aesthetics simultaneously. More importantly, the proposed deep learning framework can be easily generalized to adopt most of the existing aesthetics. Our approach is described in this section from the perspectives of training data, model architecture, loss function, and training strategy.

\subsection{Training Data and Preprocessing}
\label{sec:method-data}

As a data-driven approach, the performance of ConvGNN is impacted by the quality of the training data. A training dataset that is diverse in variety
can make the model robust to a variety of structural characteristics in the graphs,
while a homogeneous training data can inject bias into the model. Rome graphs\footnote{\url{http://www.graphdrawing.org/data.html}}, as a widely used and publicly available benchmark data set, meets our expectations. Rome contains 11,534 undirected graphs each of which consists of 10 to 100 nodes with significantly different graph structures. Therefore, our model can handle many unseen graphs regardless of their structure, thanks to a large number of graphs with different structures seen by the model during training.

Since our deep learning framework is based on ConvGNN in which the edges between nodes in the input graphs defines the flow of message passing, we made two modifications to the Rome graphs in order to facilitate the message flow within the graph. Firstly, we add virtual edges for any pairs of unconnected nodes. \xiaoqi{As mentioned in \autoref{section:background-gnn}, every node will gather information from connected nodes during convolution operation.
Thus, adding virtual edges allows information to propagate through longer distances quickly even in a shallow neural network \cite{gilmer-riley}, as messages can be directly passed between any pair of nodes. In the mean time, the original graph structure is still retained by encoding the real edge information as edge features.} Secondly, even though Rome graphs are undirected, the ConvGNN assumes its input graphs are directed. During propagation, the message flow direction is determined by the edge direction. 
%To be specific, for an edge which starts at node $u$ and ends at node $v$, the message can only propagates from $u$ to $v$. 
To avoid information asymmetry, we add duplicated edges in the opposite direction of the original edges for the connected nodes. 
%Therefore, message passing becomes two-way between nodes without hindrance and every virtually complete graph contains N(N-1) edges. 
After applying these two modifications, the Rome graphs all become complete graphs without self-loop such that there exist two edges between any pairs of nodes but in opposite directions.

The node features of graphs can also affect the learned aggregator functions. \xiaoqi{The input node features $\widetilde{\mathbf{X}}$ in (\ref{gnn-prop}) is defined as the initial node position, which allows the model to find high-quality layouts easier given a reasonable starting point as a hint.} We employ two initialization strategies. The first strategy is randomly sampling $\widetilde{\mathbf{X}} \in \mathbb{R}^{N \times 2}$ from a uniform distribution in $[0, 1]^2$. The second strategy is initializing $\widetilde{\mathbf{X}}$ as PivotMDS\cite{PMDS} layouts. 

The edge features play a key role of message passing in our model. The propagation rule \cite{gilmer-riley} in our work aggregates messages from each node's neighbors based on the information passed from the connecting edges.  Accordingly, we encode the information about the original graph structure in an edge feature matrix $\mathbf{Z} \in \mathbb{R}^{N(N-1) \times 1}$ because the structural information of the original graphs is not represented in the adjacency matrix $\mathbf{A}$. For an edge between node $u$ and node $v$, the edge attribute is specified as the graph-theoretic distance $d_{uv}$.

% which is the length of the shortest path between $u$ and $v$ in the original graph. As a result, the structural characteristic of the original graph is accessible to the aggregator function.

\begin{figure}[H]
  \begin{center}
    \adjustbox{width=\linewidth, trim={0\width} {0\height} {0\width} {0\height},clip}{\includegraphics[]{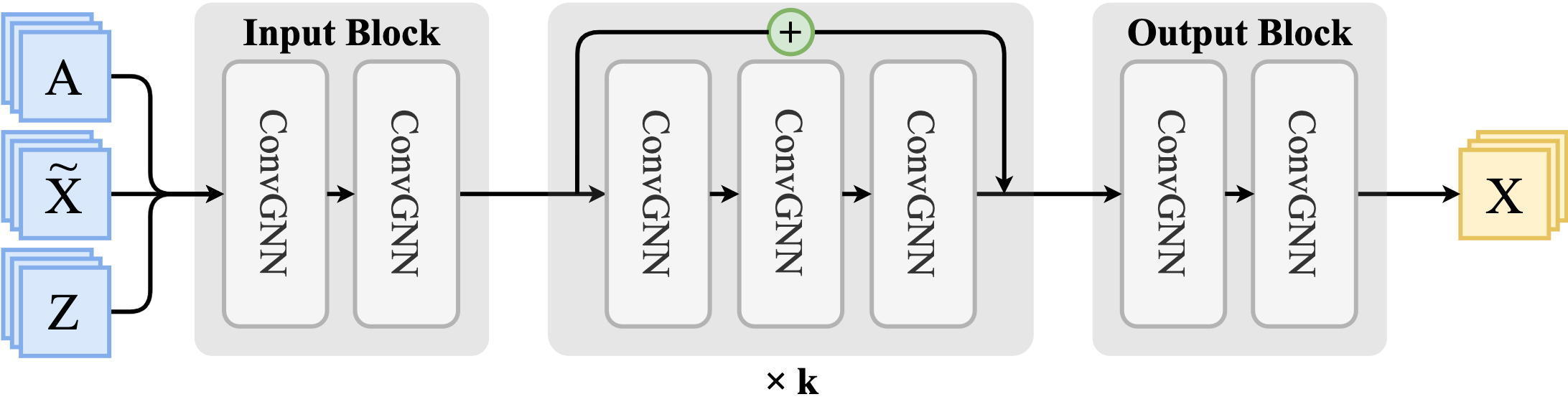}}
  \end{center}
  \caption{The model architecture of DeepGD.}
  \label{model-architecture}
\end{figure}

\subsection{Model Architecture}
The model we propose is a ConvGNN-based deep learning framework which can generate graph layouts complying with multiple aesthetic criteria simultaneously. The high-level idea is that with the graph structural characteristic captured by the convolution operation, the model predicts the node positions such that the resulted layout follows the aesthetic criteria specified by the loss function.

The input to our model includes an adjacency matrix $\mathbf{A}$, a node feature matrix $\widetilde{\mathbf{X}}$, and an edge feature matrix $\mathbf{Z}$. Overall, the model is composed of an input block, a sequence of residual blocks, and an output block (see \autoref{model-architecture}). To be specific, the input block processes and transforms the input data; a sequence of residual blocks is the key component for generating meaningful hidden node representations; and the output block is responsible for projecting the hidden node representation generated by the last residual block to two-dimensional space. 
\subsubsection{Propagation Rule} 
As stated in \autoref{section:background-gnn}, the propagation rule defines the way of aggregating messages from local neighborhood of a node. Since the original graph structure information is only maintained in the edge feature matrix $\mathbf{Z}$, our aggregator function should take into account the corresponding edge feature %(e.g., the original length of shortest path between two nodes) 
while aggregating messages. Therefore, we incorporate an edge feature network into the aggregator function such that the message passed between nodes also depends on their corresponding edge connection. This type of aggregation function is originally proposed by Gilmer et al. \cite{gilmer-riley}.
\begin{algorithm}[htbp!]
    \caption{Message Aggregation}
    \label{algorithm:aggregator}
    \SetAlgoLined
    \SetKwInput{KwInput}{Input}                % Set the Input
    \SetKwInput{KwOutput}{Output}       
    \KwInput{Graph $G(V,E)$; node embeddings $\mathbf{h} = \{\mathbf{h}_v | v \in V\}$; edge features $\mathbf{e} = \{\mathbf{e}_{uv} | (u, v) \in E\}$; weight matrix $\mathbf{W}$; edge feature network $\phi$}
    \KwOutput{Updated node embeddings $\{\tilde{\mathbf{h}}_v | v \in V\}$}
    \For{$v \in V$} {
        $\mathbf{m}_v \leftarrow \frac{1}{|\mathcal{N}(v)|}\sum_{u \in \mathcal{N}(v)} \phi \big(\mathbf{e}_{uv}) \cdot \mathbf{h}_u$ \\
        $\tilde{\mathbf{h}}_v\leftarrow \mathbf{W} \cdot \mathbf{h}_v + \mathbf{m}_v$
    }
\end{algorithm}

Our propagation rule is described in Algorithm \ref{algorithm:aggregator}. For line 2, $\mathbf{m}_v$ carries the messages aggregated from $v$'s neighbors in a complete graph $G$. Specifically, the message $\mathbf{h}_u$ from each neighbor $u$ is weighted by a transformation matrix which is generated by a learned edge feature network $\phi$ according to the corresponding edge information. As a result, this allows the model to judge the importance of the message from a specific node $u$ even if $u$ is not the close neighbor of $v$ in the original graph. After weighting the messages from all neighbors, the message is aggregated and normalized by the number of neighbors. %Since G is a complete graph, we basically normalize the total messages by $|V| - 1$. 
Then, the aggregated message $\mathbf{m}_v$ is used to update the hidden node embedding $\tilde{\mathbf{h}}_v$ on line 3. The entire message aggregation function represents the mathematical operation of one ConvGNN layer. %In other word, for each layer, all the nodes will aggregate information from their immediate neighbors once.

\subsubsection{Edge Feature Network}

In order to take the edge information into consideration during message aggregation, we design an edge feature network $\phi$ to process the edge information. We train a separate edge feature network for each ConvGNN layer because the edge information should be treated differently depending on the depth of that ConvGNN layer. The edge feature network is shared among all nodes within a single ConvGNN layer.
% \begin{figure}[ht]
%  \centering % avoid the use of \begin{center}...\end{center} and use \centering instead (more compact)
%  \includegraphics[width=\columnwidth]{pictures/edge-net.png}
%  \caption{The architecture of edge feature network.}
%  \label{edge-net}
% \end{figure}

Specifically, our edge feature network $\phi$ comprises of two dense layers. It predicts a projection matrix $\mathbf{T}_{uv} \in (-1, 1)^{F^i \times F^{i+1}}$ based on the edge feature vector $\mathbf{z}_{uv}$, where $F^i$ represents the node feature length in layer $i$. Thus, the projection matrix $\mathbf{T}_{uv}$ serves as a weighting factor for neighbor's message. Additionally, $tanh$ is used as the output activation function to confine each element in $\mathbf{T}_{uv}$ inside the range between $-1$ and $1$. 
% This aims to stabilize the message aggregation operation by preventing $T_{uv}$ from having large eigenvalues (or singular values for non-square $T_{uv}$.)

\subsubsection{Residual Blocks} 
\label{sec:residual-blocks}
The backbone of our model is composed by a sequence of residual blocks which is the key component for generating hidden node embedding. Since the graph convolution layer at a shallower depth of the network can capture low-level features and the layer at a deeper depth can learn a higher-level node representation, the residual connection can help to combine different levels of information together. 

Specifically, each residual block contains 3 hidden layers, and the input node representation of $(k-1)^{th}$ residual block will be directly added to the input node representation of $k^{th}$ residual block (see \autoref{model-architecture}). Thus, the node representation at different levels can directly pass through the model pipeline without hindrance. Moreover, skip connections facilitate the back-propagation of gradients 
so that vanishing gradient problem is alleviated.

Inspired by the stress majorization approach~\cite{gansner_stress_major}, we add two additional input edge features for each residual block. In the stress majorization algorithm (see (\ref{stress-iter})), node positions are iteratively updated by taking into account the directions of edges between each pair of node from last iteration. The reason being, the direction of edges during iteration is an important information for minimizing stress. Therefore, in addition to the original edge feature $\mathbf{d}_{uv}$, we add the direction of edges $\frac{\mathbf{h}_u-\mathbf{h}_v}{\|\mathbf{h}_u-\mathbf{h}_v\|}$ and the Euclidean distance $\|\mathbf{h}_u-\mathbf{h}_v\|$ between each pair of nodes as two additional edge features for each residual block. These two additional edge features for block $k^{th}$ are calculated according to the hidden node representation output from residual block $(k-1)^{th}$. In this case, if we regard each residual block as one iteration in (\ref{stress-iter}), each block attempts to minimize the loss function by taking the direction and edge length from last block into consideration. The entire model architecture including the residual blocks with additional edge features is described in Algorithm \ref{algorithm:convgnn}.

% \begin{figure}
\begin{algorithm*}[htbp!]
    \caption{DeepGD}
    \label{algorithm:convgnn}
    \SetAlgoLined
    \SetKwInput{KwInput}{Input}                % Set the Input
    \SetKwInput{KwOutput}{Output}       
    \KwInput{Graph $G(V,E)$; initial node features $\tilde{\mathbf{x}} = \{\tilde{\mathbf{x}}_v | v \in V\}$; edge features (graph theoretical distances) $\mathbf{d} = \{d_{uv} | (u,v) \in E\}$; total number of blocks $B$; number of layers in each block L; weight matrices $\mathbf{W}^{(b,l)}$; edge feature network $\phi^{(b,l)}$, $\forall b \in \{1,...,B\}, \forall l \in \{1,...,L\}$.}
    \KwOutput{Node positions $\mathbf{x} = \{\mathbf{x}_v | v\in V\}$}
    
    $\mathbf{h}^{(1,0)} \leftarrow \tilde{\mathbf{x}}$ \\
    $\mathbf{h}^{(1,1)} \leftarrow \text{ReLU}(\text{Message Aggregation}(G, \mathbf{h}^{(1,0)}, \mathbf{d}, \mathbf{W}^{(1,1)}, \phi^{(1,1)}))$ \\
    $\mathbf{h}^{(1,2)} \leftarrow \text{ReLU}(\text{Message Aggregation}(G, \mathbf{h}^{(1,1)}, \mathbf{d}, \mathbf{W}^{(1,2)}, \phi^{(1,2)}))$ \\
    $\mathbf{h}^{(2,0)} \leftarrow \mathbf{h}^{(1,2)}$ \\
    \For{$b \leftarrow 2...B-1$} {
        $\mathbf{e}^{(b)} \leftarrow \big\{\big(d_{uv}, \text{direction}(\mathbf{h}_u^{(b,0)}, \mathbf{h}_v^{(b,0)}), \text{distance}(\mathbf{h}_u^{(b,0)}, \mathbf{h}_v^{(b,0)})\big) \big\}$ \\
        \For{$l \leftarrow 1...L$} {
            $\mathbf{h}^{(b,l)} \leftarrow \text{ReLU}(\text{Message Aggregation}(G, \mathbf{h}^{(b,l-1)}, \mathbf{e}^{(b)}, \mathbf{W}^{(b,l)}, \phi^{(b,l)}))$
        }
        $\mathbf{h}^{(b+1,0)} \leftarrow \mathbf{h}^{(b,L)} + \mathbf{h}^{(b,0)}$
    }
    $\mathbf{h}^{(B,1)} \leftarrow \text{ReLU}(\text{Message Aggregation}(G, \mathbf{h}^{(B,0)}, \mathbf{d}, \mathbf{W}^{(B,1)}, \phi^{(B,1)}))$ \\
    $\mathbf{h}^{(B,2)} \leftarrow \text{Message Aggregation}(G, \mathbf{h}^{(B,1)}, \mathbf{d}, \mathbf{W}^{(B,2)}, \phi^{(B,2)})$ \\
    $\mathbf{x} \leftarrow \mathbf{h}^{(B,2)}$
\end{algorithm*}

\subsection{Loss Function}
\label{sec:loss}
Our loss function design largely depends on the desired aesthetics. That is, the desired aesthetics are specified in the loss function such that minimizing the loss function will thus optimize the aesthetic metrics. This enables our deep learning framework to generalize to most of the known aesthetic metrics. 

For the loss function, we have explored to optimize stress in (\ref{stress}), t-SNE in (\ref{equ:tsne-loss}), and three other commonly agreed aesthetic metrics including minimum angle, edge length variation and node occlusion.

%component shown below as part of our loss function.

%\begin{align}
%L_\text{stress}=
%\sum_{u, v \in V, u \neq v}
%\frac{
%    (l_{uv} - d_{uv})^2
%}{
%    d_{uv}^2
%},
%\qquad  l_{uv} = \|p_u - p_v\|
%\end{align}

\paragraph{\textbf{Minimum Angle}} 
%As mentioned in \autoref{section:aesthetic-criteria}, the maximal value of minimum angle is achieved when each incident angle over node $v$ equals to the desired value $\frac{2\pi}{deg(v)}$. In other word, minimizing the deviation between $\frac{2\pi}{deg(v)}$ and all incident angles over node $v$ should achieve the same goal as maximizing the minimum angle. However, 
% If we directly maximize the minimum angle metric (see \autoref{equ:minimum-angle}), the non-smooth nature of that function will make the optimization much harder. Thus, we redesign the loss function as minimizing the difference between the actual angle and the desired angle for each incident angle over node $v$. Also, we explored to use L1 loss and L2 loss to evaluate that difference.

% \begin{align}
%     L_\text{angle} = 
%     \sum_{v \in V}
%     \sum_{\theta_v^{(i)}\in \text{angles}(v)} \bigg|\frac{2\pi}{\deg(v)}-\theta_v^{(i)}\bigg|^l,
% \end{align}
% where $l \in \{1, 2\}$.

The minimum angle is the sharpest angle formed by any two edges that meet at a common vertex of the drawing\cite{purchase-helen-angle}. If a node $v$'s minimum angle is maximized, all incident edges at node $v$ will form the same angle around node $v$. Therefore, maximizing the minimum angle can help to generate aesthetically pleasing layouts. Our loss for maximizing minimum angle is computed as 
\begin{align}
\label{equ:minimum-angle}
    L_\text{angle} = 
    \sum_{v \in V}
    \sum_{\theta_v^{(i)}\in \text{angles}(v)} \bigg|\frac{2\pi}{\deg(v)}-\theta_v^{(i)}\bigg|,
\end{align}
where angle($v$) is the incident angles over node $v$ and deg($v$) denotes the node degree of $v$.

\paragraph{\textbf{Edge Length Variation}} The edge length variation is the standard deviation of edge length for all edges in a graph\cite{haleem-huamin}. If it is minimized for a graph, the length of all edges in the graph tends to be equal. From the aesthetic perspective, a graph layout with a smaller edge length variation is always preferable. The loss function for minimizing edge length variation is defined as, 
\begin{align}
L_\text{edge} = 
    \frac{1}{|E|}
    \sum_{(u,v) \in E} \frac{(l_{uv} - \bar{l})^2}{\bar{l}^2},
\qquad \begin{cases}
    l_{uv} = \|\mathbf{x}_u - \mathbf{x}_v\|\\
    \bar{l} = 1\\
\end{cases}
\end{align}
where $l_{uv}$ denotes the edge length between $u$ and v, and $\bar{l}$ is the expected edge length.

\paragraph{\textbf{Node Occlusion}} Node occlusion or node overlapping measures how densely the nodes are clustered. The global structure of graph is clearer with smaller node occlusion.
Inspired by Haleem et al. \cite{haleem-huamin}, who defines node occlusion as the total pairs of nodes closer than a threshold, we design a smooth version of node occlusion by replacing hard threshold function with a exponential function as follows,
\begin{align}\label{equation:node-occlusion}
L_\text{node occlusion} =
\sum_{u,v \in V, u \neq v}
e^{-\|\mathbf{x}_u - \mathbf{x}_v\|}.
\end{align}

\subsection{Multi-objective Training Strategy}
\label{sec:method-strategy}
In order to consider multiple aesthetic criteria at the same time, we compute a weighted sum of loss components derived from corresponding aesthetic metrics as our multi-objective loss function. The multi-objective loss function for epoch $t$ is defined as
\begin{align}\label{equation:weighted-loss}
    L^{(t)} = \sum_{k=1}^{n}\alpha_k^{(t)}L_k^{(t)},
\end{align}
where $\alpha_k^{(t)}$ ($\sum_k \alpha_k^{(t)} = 1$) represents the weight factor for $k^{th}$ component and $L_k^{(t)}$ is the average loss value for the $k^{th}$ component at epoch $t$.
 
Different combinations of weight factors will result in different optimization results. Intuitively, the weight factors should be specified based on human preferences. In other words, if $\alpha_1$ is greater than $\alpha_2$, more emphasize is put on optimizing the first loss component. However, the weight factors do not always  directly reflect the human preference of each loss component during optimization, because different loss components may have  different numerical scales. So, one challenge is how to determine the weight factor $\alpha$ for each aesthetic while considering both the human preference and the difference in the numerical scale.

We discuss two multi-objective 
training strategies to help find a compromise between multiple aesthetics in the following.

\subsubsection{Adaptive Weight}
\label{sec:importance}
Adaptive weight is to adaptively adjust $\alpha$ with respect to the numerical scale of corresponding loss component for each epoch. The weight factor for $k^{th}$ components at epoch $t$ is defined as
\begin{align}\label{equation:adaptive-importance}
    \alpha_k^{(t)} = \frac{\frac{\gamma_k}{L_k^{(t-1)}}}{\sum_{l=1}^n \frac{\gamma_l}{L_l^{(t-1)}} ,
}
\end{align}
where $\gamma_k$ represents the importance factor for $k^{th}$ component and $L_k^{(t-1)}$  the average loss value for $k^{th}$ component in epoch $t-1$. 

The importance factor basically indicates the human preference for each loss component without considering the difference between numerical scale for multiple loss components. The basic idea is that, we normalize each component by its numerical scale $L_k^{(t)}\big/L_k^{(t-1)}$ and then multiply it with their user-specified human preference $\gamma_k$. In this case, the weight factor $\alpha_k^{(t)}$ will be decreased if the numerical scale of $k^{th}$ component is greater than that of other components, and vice versa. Thus, the weight factor for each epoch takes into account both the human preference specified as importance factor and the numerical scale of all loss components from previous epoch.

\subsubsection{SoftAdapt by Importance}
The SoftAdapt by Importance strategy is derived from the SoftAdapt technique proposed by Heydari et al. \cite{softadapt}. SoftAdapt adaptively sets the weight factor for each loss component according to their recent rate of descent. Since the original SoftAdapt does not consider different numerical scales for loss components, we improve it by taking into account the relative scale of each loss components and incorporating the concept of importance factor $\gamma_k$. Specifically, the weight factor $\alpha_k^{(t)}$ for $k^{th}$ loss component during epoch $t$ is computed as,
\begin{align}
    \alpha_k^{(t)} & = 
\frac{
    \frac{\gamma_k}{L_k^{(t-1)}} \exp(\beta^* s_k^{(t)})
}{
    \sum_{l=1}^n \frac{\gamma_l}{L_l^{(t-1)}} \exp(\beta^* s_l^{(t)})
},
\\
    s_k^{(t)} & = 
    \underset{k \in \{1 \ldots n\}}{\text{normalize}_{L1}} \bigg(\frac{
        \tilde{L}_k^{(t)} - \tilde{L}_k^{(t-1)}
    }{
        \tilde{L}_k^{(t-1)}
    }\bigg),
\\
\tilde{L}^{(t)} &=
\begin{cases}
    L^{(1)} & t = 1 \\
    \tau \cdot \tilde{L}^{(t-1)} + (1 - \tau) \cdot L^{(t)}  & t > 1, \\
\end{cases}
\end{align}
where $\beta$ is the sensitivity factor that controls how responsive the weight adjustment is to the rate of descent; $\tilde{L}_k$ is the smoothed version of $L_k$ after exponential smoothing with smoothing factor $\tau$; and $s_k^{(t)}$ is the descending rate of $k^{th}$ loss component at epoch $t$.

% To interpret this training strategy, it can be decomposed into two components: adaptive importance $\gamma_l/L_l^{(t-1)}$ and SoftAdapt $e^{\beta^* s_k^{(t)}}$. The sensitivity factor $\beta$ defines the behavior of exploitation and exploration. To be more clear, the optimizer will explore for optimizing a loss component with bad performance in the past when $\beta>0$ and exploit the loss component with the best performance when $\beta<0$.

% Compared with the original SoftAdapt, it sets a deterministic $\beta$ value so that it eliminates the potential improvement by switching the behavior between exploration and exploitation. Therefore, the second and last modification we made to the original SoftAdapt is borrowing the concept of exploration and exploitation from reinforcement learning. Specifically, with pre-specified exploitation rate $q$, we will exploit the best performance loss component for $q\%$ of times and explore the worse performance component for $(1-q)\%$ of times. For each epoch, a die will be rolled to determine the behavior of exploitation or exploration. In this case, we can make a compromise between exploitation and exploration instead of setting a deterministic $\beta$ value.

\section{Evaluation}
We conducted extensive experiments to evaluate our approach quantitatively and qualitatively. This section describes the experimental details and evaluation results.

% Firstly, the details of experimental setting will be introduced in \autoref{sec:experiment-setup} and \autoref{sec:model-config}. Secondly, the quantitative and qualitative evaluation of our approach is presented in \autoref{sec:stress-result} and \autoref{sec:composite-result}. Thirdly, the comparative study of model architecture and
% multi-objective training strategy is conducted in \autoref{sec:model-archi-result} and 
% \autoref{sec:strategy-result}. Lastly, the computational performance is evaluated in \autoref{sec:train-test-time}. 

\subsection{Experimental Setup}
\label{sec:experiment-setup}
The proposed deep learning framework is implemented with PyTorch and PyTorch Geometric library\footnote{\url{https://github.com/rusty1s/PyTorch_geometric}}. Besides, in all of our experiments, the models were trained on a single Tesla V100 GPU with memory of 16 GB.

As mentioned in \autoref{sec:method-data}, our experiments are conducted over Rome dataset, which contains 11,534 undirected graphs each consisting of 10 to 100 nodes. We excluded two disconnected graphs from the dataset. Among the remaining graphs, we randomly selected 10,000 graphs as training examples, 1,000 graphs as testing examples, and 532 graphs as validation examples. Validation data serve the purpose of hyperparameter tuning. We only report the model performance on testing data in this paper.

\xiaoqi{Regarding model architecture, we conducted a series of experiments/ablation studies to investigate how does each component contribute to the model performance. Those experiments explored the effect of removing residual connection, removing all the virtual edges, removing different edge features, different numbers of neurons for each residual block, and different numbers of hidden layers in the edge feature network. Given the word limits, our model configuration and detailed experimental results for comparing different model architecture are presented in \hyperref[appendices]{Appendices}}.

% In addition to the original Rome graphs, we augmented it with randomly modified nodes and edges. To be specific, to modify nodes, we randomly add $x_1 \thicksim Uniform(1, \frac{N}{4})$  numbers of nodes and remove $x_2\thicksim Uniform(1, \frac{N}{4})$ number of nodes for each Rome graph. After modifying nodes, there are $x_3 \thicksim Uniform(1, \frac{|E|}{4})$ edges to be randomly removed and $x_4 \thicksim Uniform(1, \frac{|E|}{4})$ edges to be randomly added. Hence, we generated two augmented data sets containing 10,000 training graphs based on this augmentation strategy. For one of augmented data, we only modified edge connections in the graph. However, for another augmented data, both the nodes and edges are randomly modified. To explore the benefit of larger training data size, the model performance is also evaluated on these two augmented data sets.

\subsection{Quantitative Evaluation}
In this section, we quantitatively assess the performance of DeepGD. For comparison, PivotMDS\cite{PMDS} and the stress majorization\cite{gansner_stress_major} algorithm $neato$ implemented in Graphviz\footnote{\url{https://graphviz.org/}} are chosen as baseline methods.

% We like to know if by using a dimension reduction algorithm to ``project'' the high dimensional output of the last hidden layer of the GNN might give good results, to that end we implemented GNN+tSNE and GNN+UMAP. These two baselines used t-SNE and UMAP, respectively, to project the graph embedding generated by the last hidden layer of DeepGD into 2-D space.

In addition to Graphviz and PivotMDS, inspired by Tsitsulin et al.\cite{VERSE} who propose to use t-SNE to project high dimensional  node embedding to 2D for graph visualization, we implemented GNN+tSNE and GNN+UMAP. These two baselines used t-SNE and UMAP, respectively, to project the latent node embedding generated by the last hidden layer of DeepGD onto 2-D space.

To evaluate the relative difference, we computed Symmetric Percent Change (SPC) with respect to Graphviz as follows.
\begin{equation}
    \label{equ:spc}
    \text{SPC} = 100\% \times \frac{1}{N_t}\sum_{i=0}^{N_t}\frac{D_i-G_i}{max(D_i,G_i)}, \\
\end{equation}
% where $D_i$ indicates the metric of graph layouts generated by DeepGD for the $i^{th}$ test graph; $G_i$ represents the metric of graph layouts drawn by Graphviz for the $i^{th}$ test graph; and $N_t$ denotes the total number of test graphs.
where $D_i$ and $G_i$ denotes the same evaluation metric computed on two layouts generated by a certain model (i.e., DeepGD or PivotMDS) and Graphviz respectively, for the $i^{th}$ test graph; $N_t$ is the total number of test graphs. SPC holds a nice property that it ranges from $-100\%$ to $100\%$, thus the value of SPC can be interpreted as how many percent $G_i$ outperforms $D_i$.

\subsubsection{Stress Optimization}
\label{sec:stress-result}
Since minimizing stress is known to be an overall effective approach to improve many aesthetic aspects of graph layouts, we first explored the effectiveness and efficiency of optimizing stress only. In other words, the loss function of DeepGD in this subsection contains only one component which corresponds to stress.

\begin{table}[htbp!]
\renewcommand{\arraystretch}{1.4}
\setlength{\tabcolsep}{6.5pt}
\caption{To assess the DeepGD with stress only, average stress is computed over 1000 test graphs; and the stress SPC (smaller is better) represents the relative difference in stress compared to Graphviz. }
\label{table:stress}
\centering
\begin{tabular}{c|c<{\centering}|c<{\centering}}
\bfseries Models & \bfseries Avg. Stress & \makecell{\bfseries Stress SPC \\ \bfseries w.r.t. Graphviz} \\
\hline
\rule{0pt}{3ex}
\bfseries{DeepGD + Random Init} & \bfseries{246.57} & 1.10\%\\
\bfseries{DeepGD + PivotMDS Init} & \bfseries{239.73}& \bfseries{-5.43\%}\\
\bfseries{Graphviz\cite{gansner_stress_major}} & 251.93 & 0.00\%\\
\bfseries{PivotMDS\cite{PMDS}} & 372.06 & 36.53\%\\
\bfseries{GNN + t-SNE\cite{tsne}} & 483.01 & 56.98\%\\
\bfseries{GNN + UMAP\cite{umap}} & 379.88 & 41.40\% \\ 
\end{tabular}
\end{table}

As shown in \autoref{table:stress}, DeepGD obtains better stress than Graphviz on average, no matter what the initialization strategy is. For stress SPC, DeepGD initialized with PivotMDS outperforms Graphviz by 5.43\% whereas DeepGD with random initialization achieves a comparable performance as Graphviz. We observe that DeepGD initialized randomly achieves a positive stress SPC but with lower stress than Graphviz. The potential reason is that DeepGD usually outperforms Graphviz for large graphs but obtains higher stress than Graphviz for drawing small graphs. Besides, if we compare the three alternative baselines (PivotMDS, GNN+t-SNE, and GNN+UMAP) with Graphviz, their performance are much worse. In particular, projecting the output of the last hidden layer using t-SNE or UMAP is inferior to DeepGD, showing the importance of end-to-end training by including the final nonlinear layer.
In conclusion, regarding stress, DeepGD significantly outperforms PivotMDS, GNN+ t-SNE, and GNN+ UMAP and is 5.43\% better than Graphviz on average.

% When we inspect the data, we observed some outliers in the test set. For example, there is a graph which can be drawn in a straight line, and the optimal stress for that graph should be very close to zero. For this graph, a small difference between DeepGD stress and Graphviz stress result in a large stress SPC because of the small denominator. %Therefore, the stress SPC might be significantly affected by those extreme cases. 
% This explains why the stress SPC is positive even though DeepGD mean stress is smaller than that of Graphviz. 

% To evaluate the stability and robustness of DeepGD, we performed a k-fold cross validation. With 11,000 Rome graphs, we split it into 11 folds. For each fold, there are 10,000 training graphs and 1,000 test graphs. 11-folds cross-validation on DeepGD with random initialization achieves stress SPC of 2.49\% on average. Additionally, the median stress SPC for every folds are only slightly above zero, which indicates that DeepGD with stress performs as good as Graphviz regardless of the outliers. We can conclude that DeepGD is able to consistently perform well even with the variation in the training data.

To evaluate the stability and robustness of DeepGD, we performed 11-fold cross validation over 11,000 Rome graphs. With random initialization, DeepGD achieves stress SPC of 1.10\% on average. Additionally, the median stress SPC for each folds are only slightly above zero, which indicates that DeepGD with stress performs as good as Graphviz regardless of the outliers. We can conclude that DeepGD is able to consistently perform well even with the variation in the training data. 

% \begin{table*}[htbp!]
%     \caption{The qualitative evaluation of DeepGD with stress only. }
%     \label{table:stress-result}
%     \fontsize{8}{12}\selectfont
%     \centering
%     \begin{tabular}{ l|>{\centering\arraybackslash}m{10em} >{\centering\arraybackslash}m{10em} }
%         \toprule
%         & {Rome\#9102} & {Rome\#2364} \\ 
%         \midrule
%         DeepGD & \includegraphics[height=7.5em]{pictures/our-10145.png} & \includegraphics[height=7.5em]{pictures/our-10786.png} \\
%         Graphviz & \includegraphics[height=7.5em]{pictures/GV-10145.png} & \includegraphics[height=7.5em]{pictures/GV-10786.png} \\ 
%         \bottomrule 
%     \end{tabular}
% \end{table*} 

\subsubsection{Optimization with Two Aesthetics}
\label{sec:composite-result}
We assess the model capability of optimizing and compromising between two aesthetics metrics in this section. Since stress can improve the overall aesthetic quality, we conducted experiments for combining stress and one other aesthetic metrics including t-SNE, edge length variation, minimum angle, and node occlusion respectively. The following three models are trained by adaptive weight strategy mentioned in \autoref{sec:importance} with our choice of importance factors.

%In addition to optimizing two aesthetics, the effectiveness of DeepGD optimizing those four aesthetic metrics simultaneously is also evaluated.

% To the best of our knowledge, there is no existing algorithm which attempts to explicitly optimize stress and those aesthetics simultaneously, therefore we do not have a strict ground truth model for optimization with multiple aesthetics to compare with. Hence, we regard the stress majorization implemented in Graphviz as the comparison baseline for stress, minimum angle, and node occlusion. 
% Besides, for t-SNE loss, the baseline model is supposed to be tsNET because it is an iterative approach for optimizing t-SNE loss. However, the tsNET implementation (\url{https://github.com/HanKruiger/tsNET}) generates random looking drawing for some of test graphs and consequently does not serve well as a baseline. Therefore, we still use stress majorization in Graphviz as the baseline for comparing t-SNE.

\begin{table*}[htbp!]
\setlength{\tabcolsep}{2.0pt}
\caption{The quantitative evaluation of DeepGD with multiple aesthetics. Each row represents one DeepGD model with our choice of aesthetics, which are weighted linear 
combinations of different loss components (stress, occlusion etc.). Negative SPC indicates that DeepGD outperforms Graphviz with certain percentage regarding that specific metric.}
\label{table:deepgd-more-aesthetics}
\centering
\begin{tabular}{c|c|c|c|c|c|c|c|c|c|c|c|c|c|c}
\multicolumn{5}{c|}{\multirow{2}{4cm}[0pt]{\bfseries\makecell{Importance Weighting\\Factors of Loss Components}}} & \multicolumn{10}{c}{\bfseries Metric SPC w.r.t. Graphviz}
\rule{0pt}{2.2ex} \\ 
\cline{6-15}

\multicolumn{5}{c|}{} & 
\multicolumn{5}{c|}{\bfseries Random Initialization} &
\multicolumn{5}{c}{\bfseries PivotMDS Initialization} 
\rule{0pt}{2.2ex} \\ 
\hline

Stress & Angle & Edge & Occlusion & t-SNE & Stress & Angle & Edge & Occlusion & t-SNE & Stress & Angle & Edge & Occlusion & t-SNE 
\rule{0pt}{2.2ex} \\
\hline

\rule{0pt}{2.4ex}
0.6 & 0.4 &     &     &     & 17.59\% & \bf{-17.10\%} &   --     &   --   &   --    &  4.24\% & \bf{-22.66\%} &   --     &   --    &   --     \rule{0pt}{2.2ex}\\
0.8 &     & 0.2 &     &     &  4.88\% &   --     & \bf{-20.28\%} &   --   &   --    &  4.64\% &   --     & \bf{-32.92\%} &   --    &   --     \rule{0pt}{2.2ex}\\
0.6 &     &     & 0.4 &     &  0.72\% &   --     &   --     & 3.01\% &   --    & \bf{-4.67\%} &   --     &   --     & \bf{-2.70\%} &   --     \rule{0pt}{2.2ex}\\
0.7 &     &     &     & 0.3 &  1.88\% &   --     &   --     &   --   & \bf{-5.18\%} & \bf{-3.84\%} &   --     &   --     &   --    & \bf{-12.09\%} \rule{0pt}{2.2ex}\\
0.5 & 0.1 &     & 0.1 & 0.3 &  4.19\% & \bf{-0.60\%}  &   --     & 0.76\% & \bf{-7.29\%} & \bf{-1.53\%} & \bf{-7.36\%}  &   --     & \bf{-2.09\%} & \bf{-14.48\%} \rule{0pt}{2.2ex}\\

\end{tabular}
\end{table*}

% model 159
\paragraph{\textbf{Stress + Minimum Angle Loss}}
In this DeepGD model, the stress and minimum angle loss are optimized at the same time. From the first row of \autoref{table:deepgd-more-aesthetics}, DeepGD with both initialization strategies outperforms Graphviz by at least 17\% considering minimum angle loss. However, the stress SPC of DeepGD with random initialization and PivotMDS initialization increases by 16.49\% and 9.67\%, respectively, compared to the DeepGD with stress only in first two rows of \autoref{table:stress}. The potential reason is that minimum angle and stress are conflicting criterion so that there is an unavoidable trade-off between them.

% We also present an example layout generated by DeepGD in the second row of \autoref{table:aesthetic-combinations}. It is very obvious that by optimizing stress and minimum angle simultaneously, DeepGD can distribute the incident angles more evenly than Graphviz. 

\paragraph{\textbf{Stress + Edge Length Variation}}
We also conducted experiment to optimize stress and edge length variation simultaneously. In the second row of \autoref{table:deepgd-more-aesthetics}, the edge length variation SPC of -20.28\% and -32.92\% shows that DeepGD can draw a graph with much more uniform edge length than Graphviz on average. It is interesting to see that DeepGD with both initialization strategies still obtains a reasonably good stress, even though it has to compromise between stress and edge length variation. This is because stress will be minimized when the layout distance between each node pairs equals to their graph theoretic distance. Therefore, minimizing edge length variation could also help with minimizing stress.

% From the third row of \autoref{table:aesthetic-combinations}, we can see that the edge length of DeepGD layout is more uniform than that of Graphviz layout.

\paragraph{\textbf{Stress + Node Occlusion}}
From the third row of \autoref{table:deepgd-more-aesthetics}, we observe that the performances of DeepGD with two initialization strategies are slightly different. With random initialization, DeepGD obtained 3.01\% higher node occlusion loss than Graphviz on average. Given that Graphviz only optimizes stress, it indicates that optimizing stress can help to avoid node occlusion as well. With PivotMDS initialization, we outperform Graphviz regards both stress and node occlusion by 4.67\% and 2.70\%, respectively. Overall, this experimental result again proves that DeepGD has the capability and flexibility of optimizing most of aesthetics.

\paragraph{\textbf{Stress + t-SNE}} To optimize stress and t-SNE simultaneously, the loss function in this DeepGD model is the weighted average of stress and t-SNE (see \autoref{sec:method-strategy}). The quantitative measurement shown in the fourth row of \autoref{table:deepgd-more-aesthetics} indicates that DeepGD with PivotMDS outperforms Graphviz regarding both t-SNE and stress. Also, with random initialization, we achieve better t-SNE and comparable stress than Graphviz. This result again shows that PivotMDS initialization indeed can help to improve the performance. 

% In addition, the qualitative evaluation presented in first row of \autoref{table:aesthetic-combinations} manifests that with t-SNE loss, DeepGD can indeed draw a good graph layout.

% Also, comparing the example layouts in the fourth row of \autoref{table:aesthetic-combinations}, the nodes in DeepGD layouts are more disperse than Graphviz layouts.

\subsubsection{Optimization with Four Aesthetics}
To assess the model's capability of optimizing more than two aesthetics, we conducted experiments for optimizing stress, t-SNE, minimum angle, and node occlusion simultaneously. The quantitative evaluation is presented in the fifth row of \autoref{table:deepgd-more-aesthetics}. By initializing DeepGD with PivotMDS, we achieved outstandingly better results than Graphviz from four different aesthetic perspectives. This result clearly demonstrates that DeepGD indeed can draw arbitrary graphs by balancing among multiple aesthetics with the help of adaptive importance training strategy, even though t-SNE and minimum angle loss somehow contradicts with stress. More importantly, since a layout method that considers multiple aesthetics simultaneously is more likely to generate a visually pleasing graph layout \cite{huang-2013}, DeepGD might be aesthetically more attractive to human.

\subsection{Qualitative Evaluation}
For qualitative evaluation shown in \autoref{fig:vis-result}, we only present the results for DeepGD with PivotMDS initialization due to the significantly better result comparing to DeepGD with random initialization. Among all 10 methods, PivotMDS, GNN+t-SNE, and GNN+UMAP each have obviously identifiable weaknesses that make them not as visually pleasing as the rest, which is consistent with the results shown in \autoref{table:stress}. In contrast, DeepGD generated layouts on all sample graphs are at least comparably good to Graphviz. We also observe that, for DeepGD models, the specific loss components involved indeed improve the corresponding visual aspects of the resulting layouts.

\newcommand{\imgcell}[1]{\adjustbox{height=7.8em, trim={0.08\width} {0.13\height} {0.08\width} {0.09\height},clip}{\includegraphics[]{#1}}}
\begin{table*}[ht!]

\setlength{\tabcolsep}{0pt}
\renewcommand{\arraystretch}{0}
    \fontsize{6}{6}\selectfont
    \centering
    \begin{tabular}{ cccc|cccccc }
        \multicolumn{4}{c|}{\thead{Baseline Methods}} & \multicolumn{6}{c}{\thead{DeepGD}}\\
         \bfseries{Graphviz} & \bfseries{PivotMDS} & \bfseries{GNN+t-SNE} & \bfseries{GNN+UMAP} & \bfseries{Stress} & \bfseries{Stress + t-SNE} & \bfseries{Stress + Angle} & \bfseries{Stress + Edge} & \bfseries{Stress + Occlusion} & \bfseries{Four Aesthetics} \rule[-1ex]{0pt}{0ex} \\ \hline
         
\imgcell{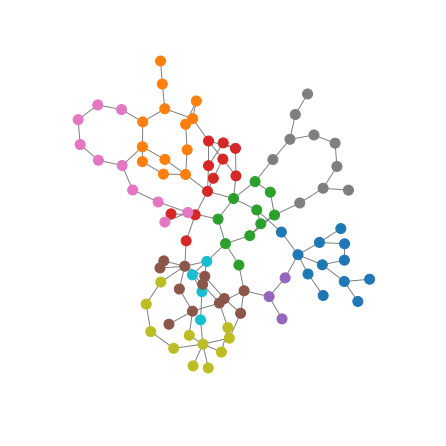}&\imgcell{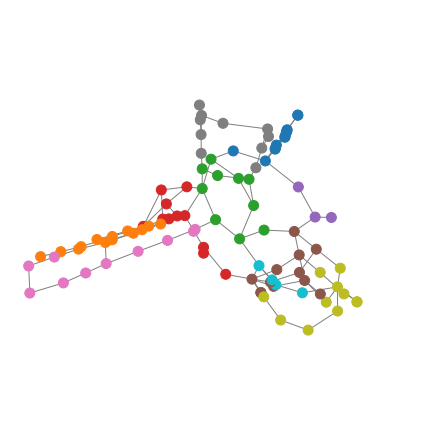} &\imgcell{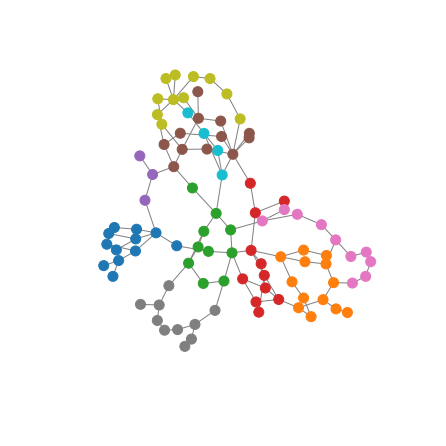}&\imgcell{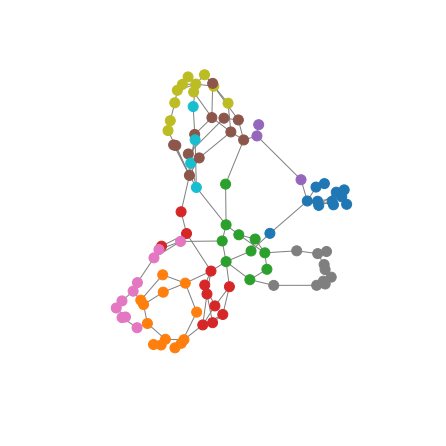}&\imgcell{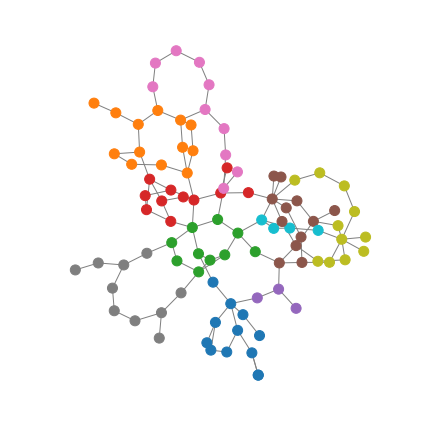}&\imgcell{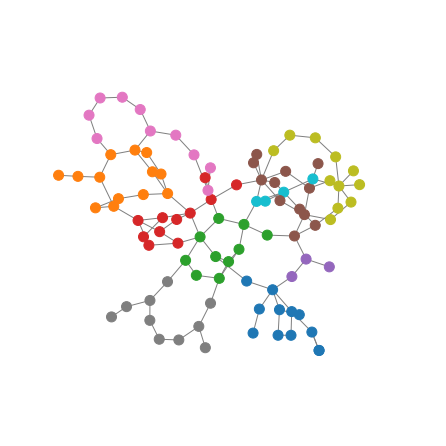}&\imgcell{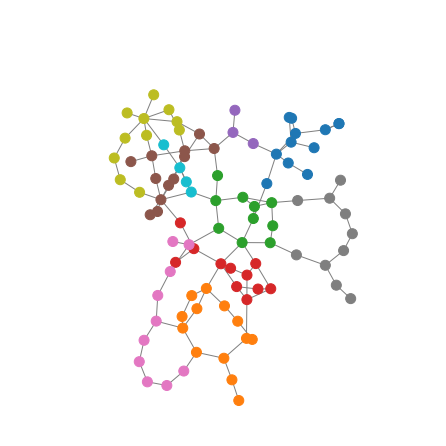}&\imgcell{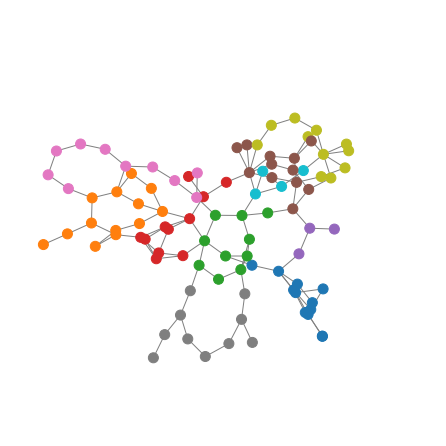}&\imgcell{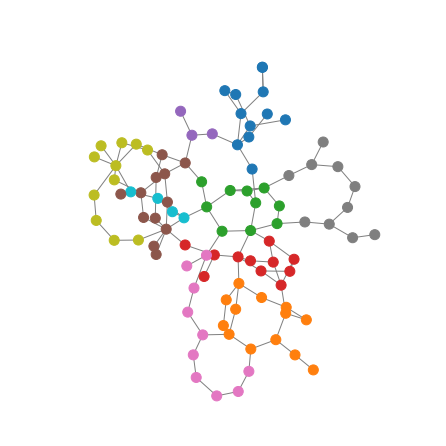}&\imgcell{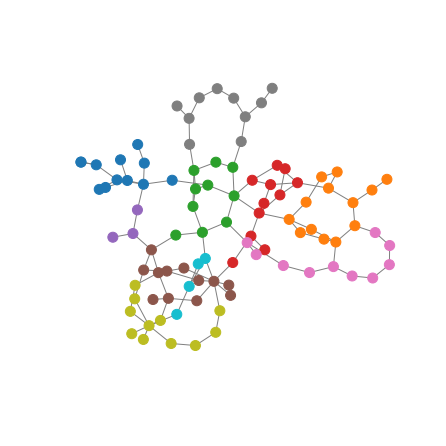}\\

\imgcell{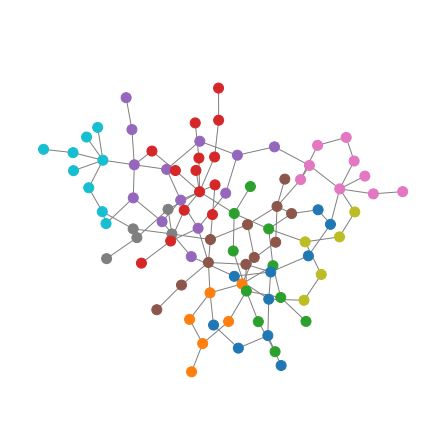}&\imgcell{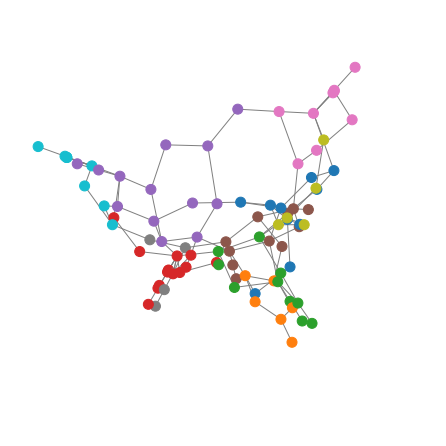} &\imgcell{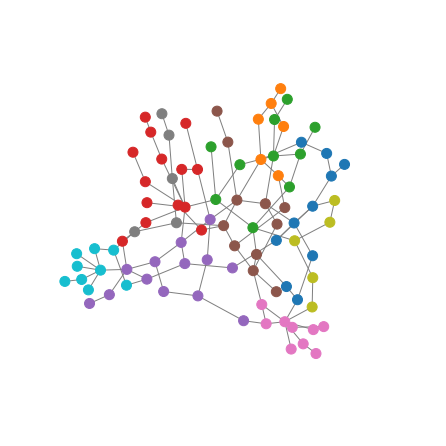}&\imgcell{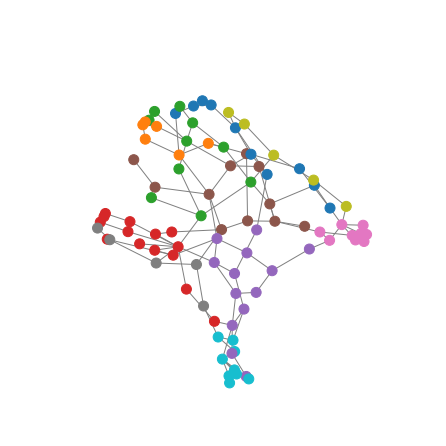}&\imgcell{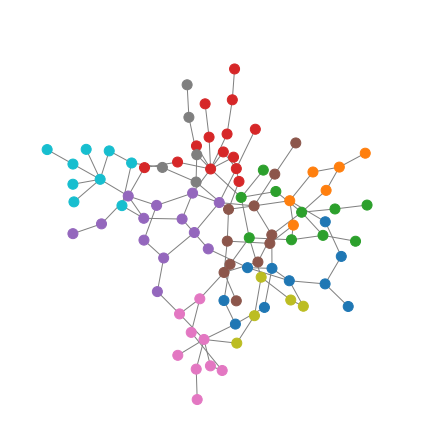}&\imgcell{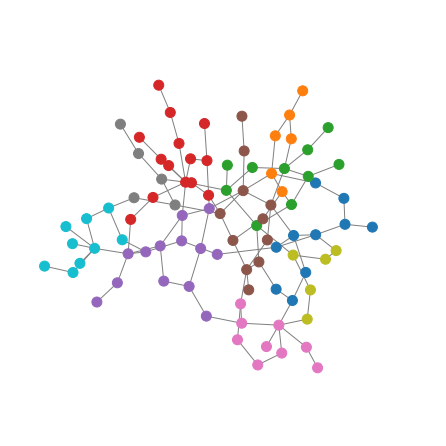}&\imgcell{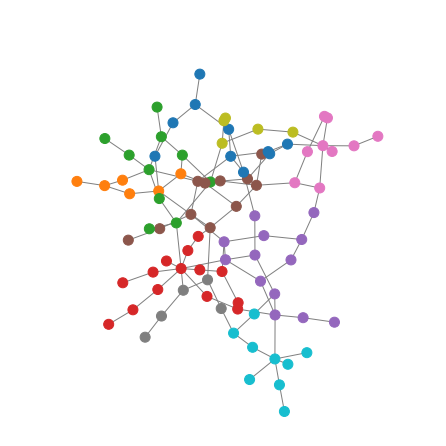}&\imgcell{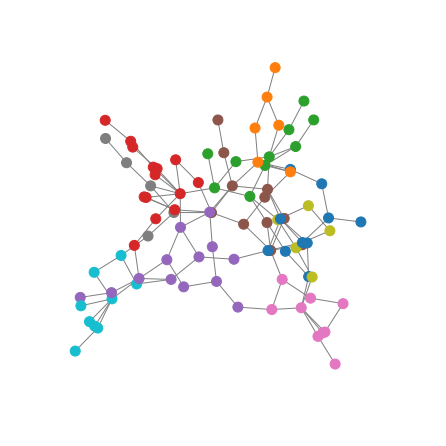}&\imgcell{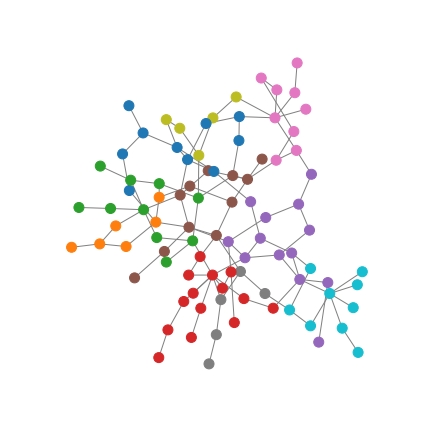}&\imgcell{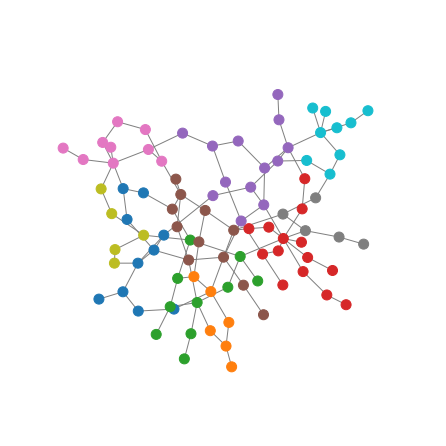}\\

\imgcell{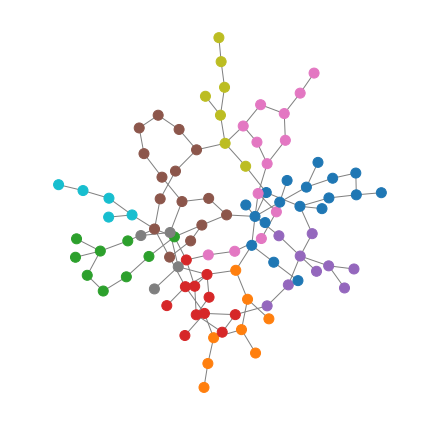}&\imgcell{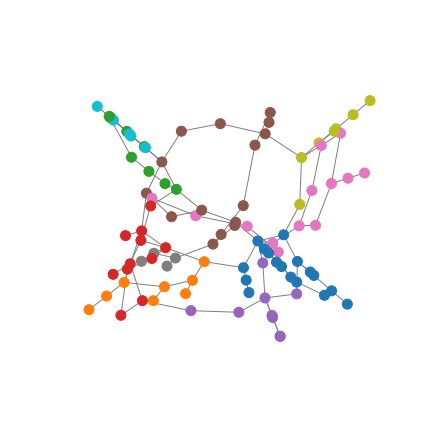} &\imgcell{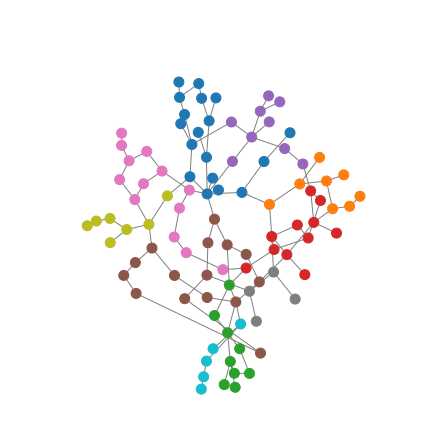}&\imgcell{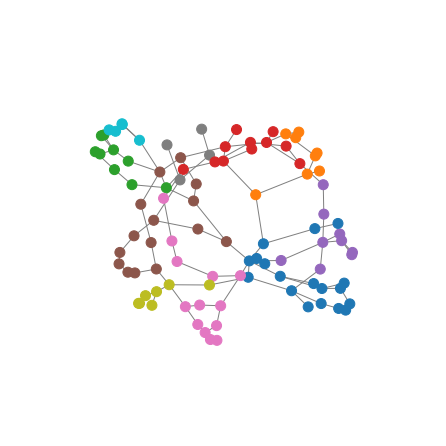}&\imgcell{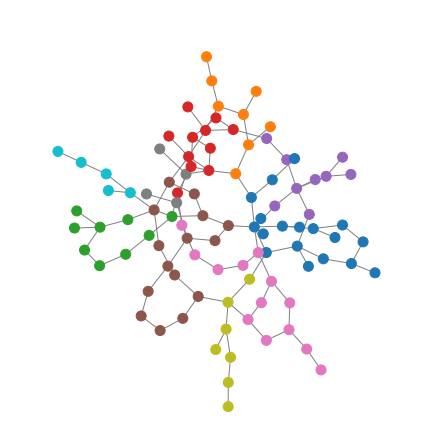}&\imgcell{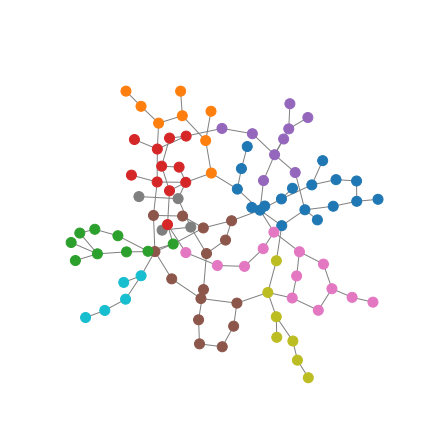}&\imgcell{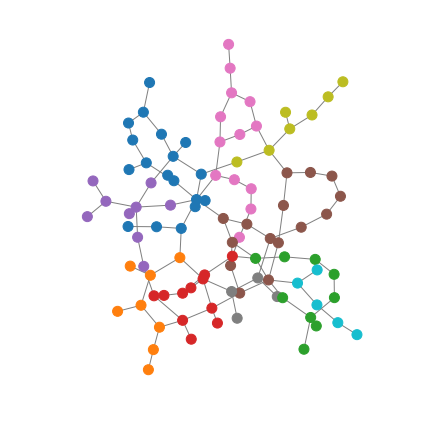}&\imgcell{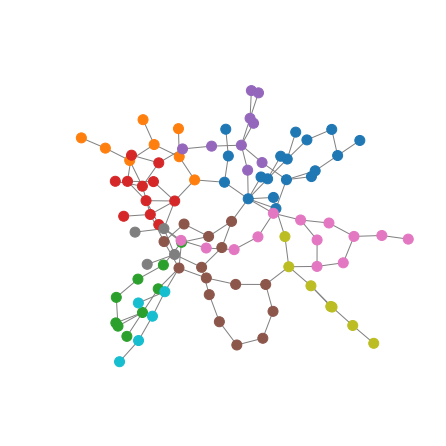}&\imgcell{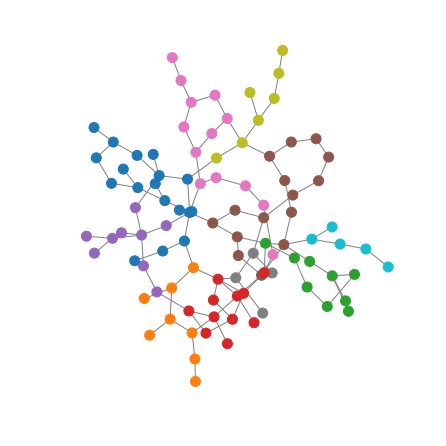}&\imgcell{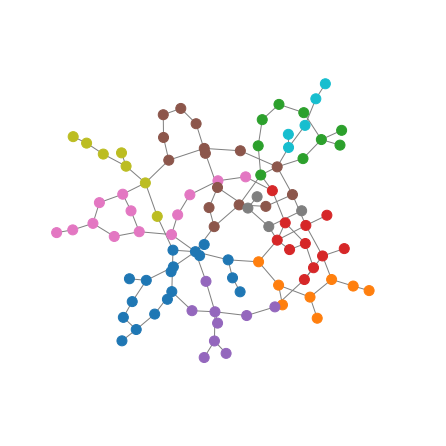}\\

\imgcell{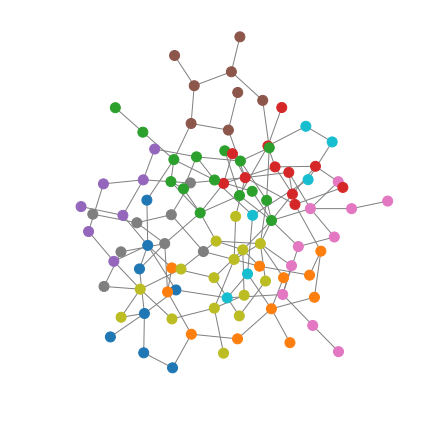}&\imgcell{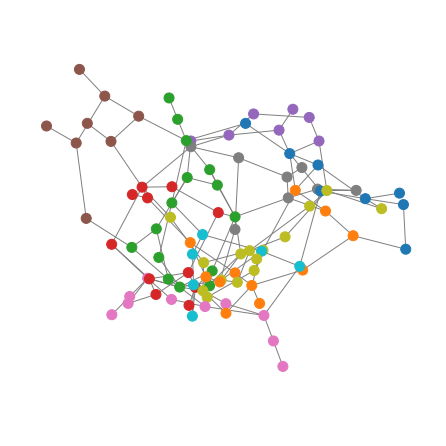} &\imgcell{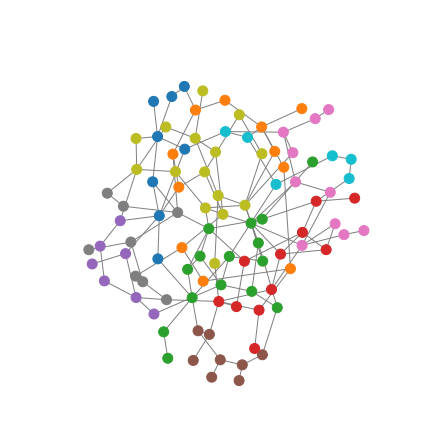}&\imgcell{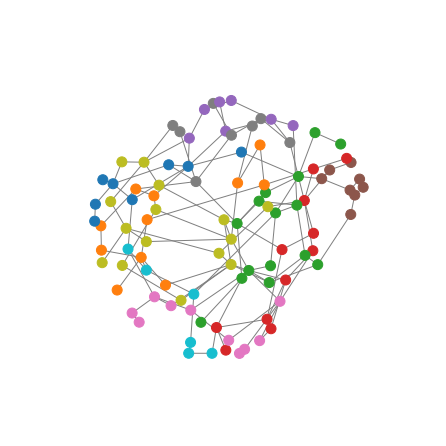}&\imgcell{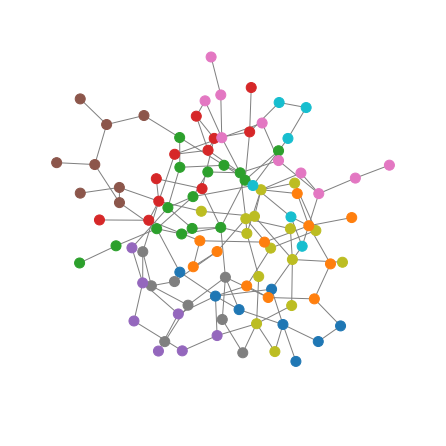}&\imgcell{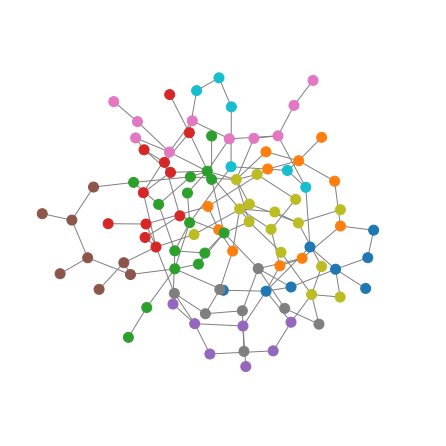}&\imgcell{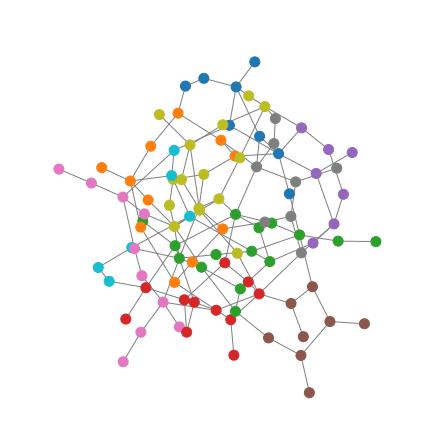}&\imgcell{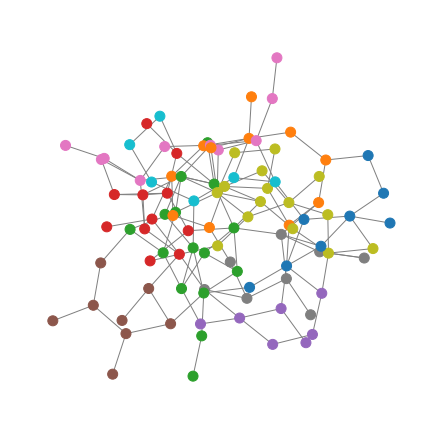}&\imgcell{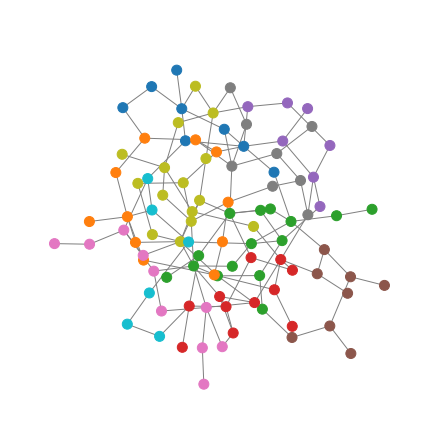}&\imgcell{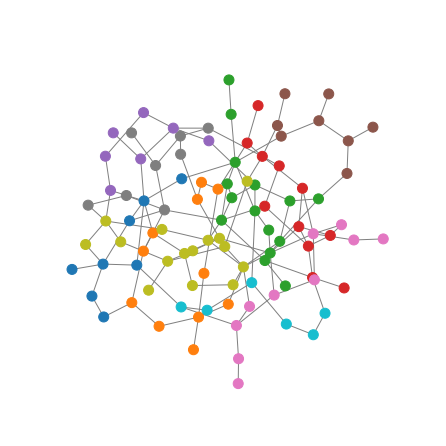}\\

\imgcell{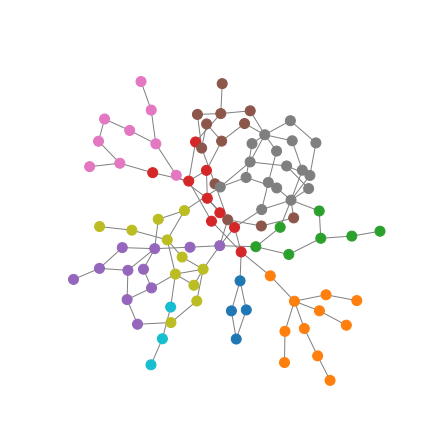}&\imgcell{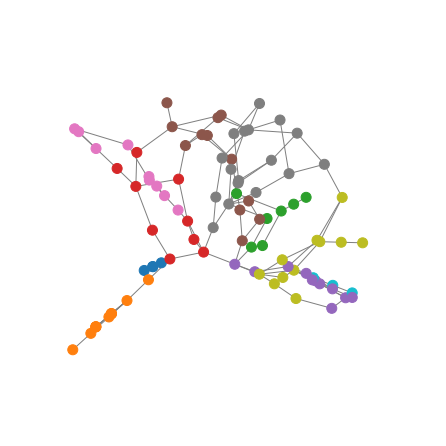} &\imgcell{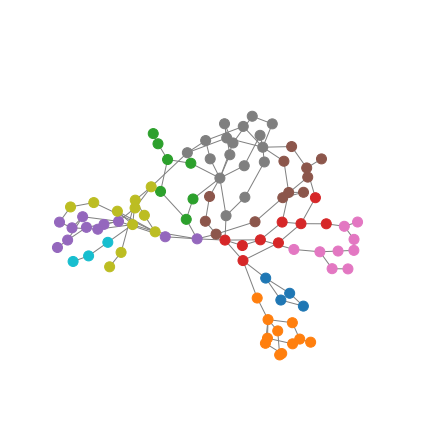}&\imgcell{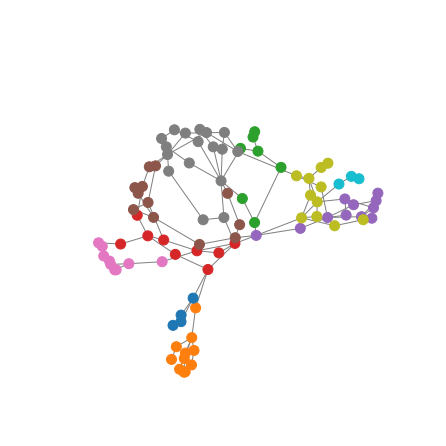}&\imgcell{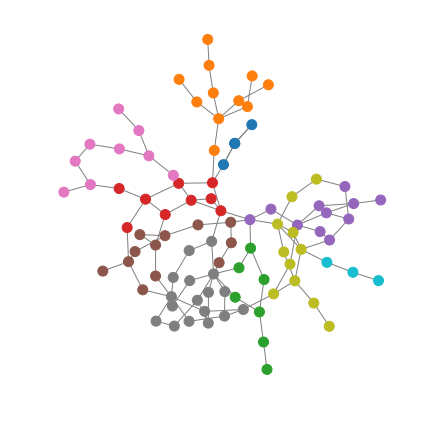}&\imgcell{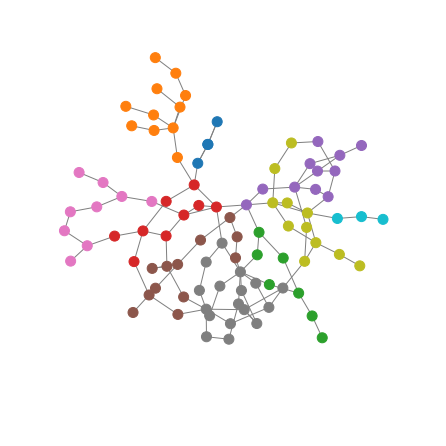}&\imgcell{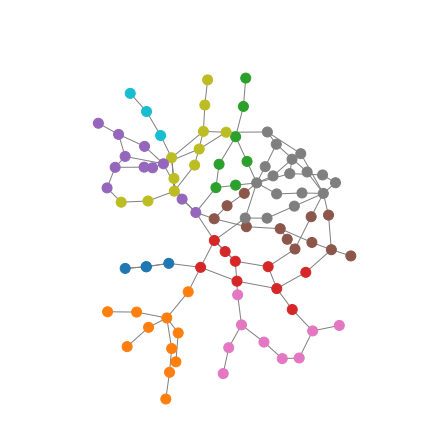}&\imgcell{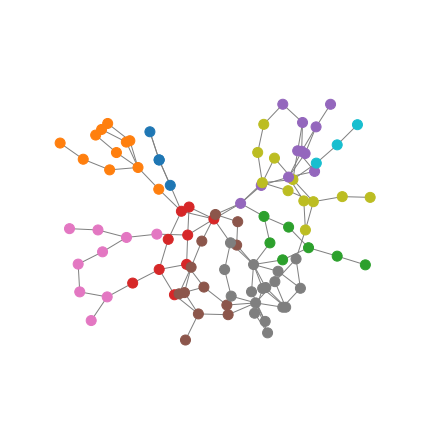}&\imgcell{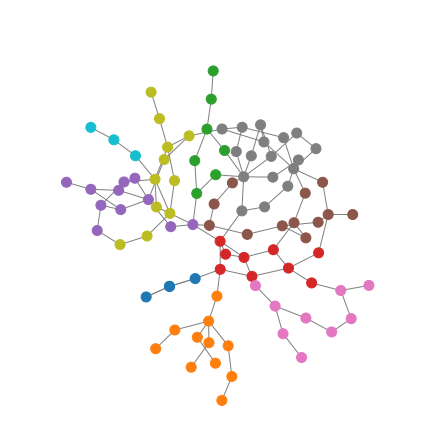}&\imgcell{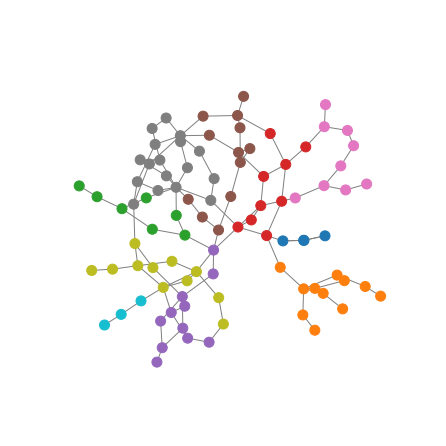}\\

\imgcell{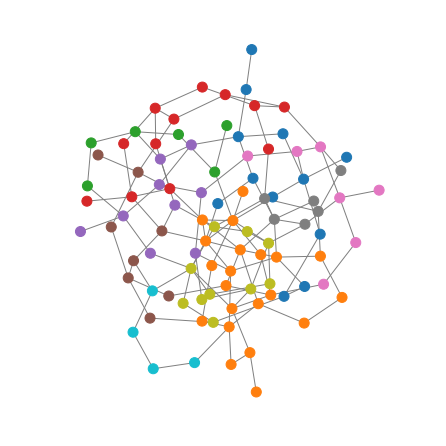}&\imgcell{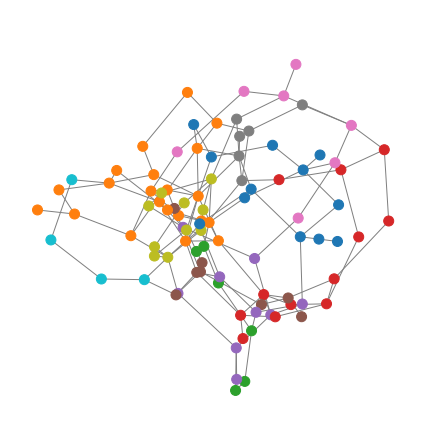} &\imgcell{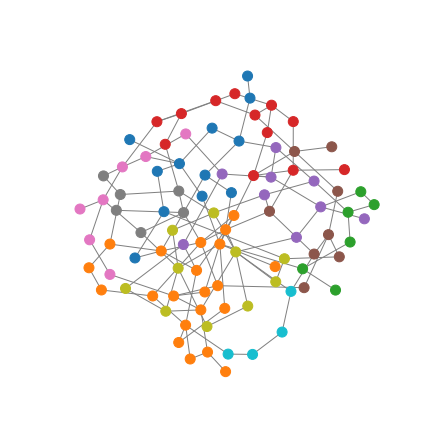}&\imgcell{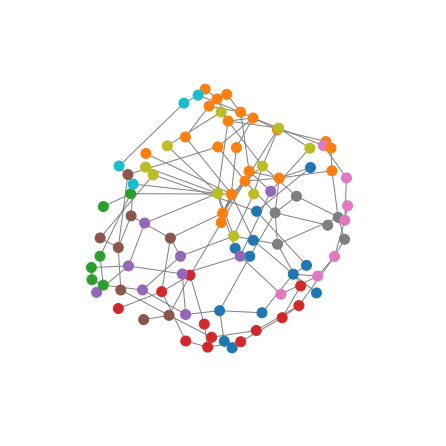}&\imgcell{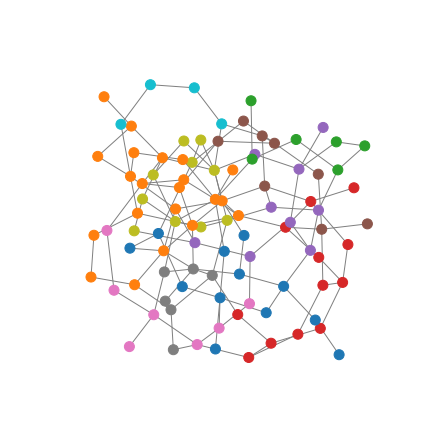}&\imgcell{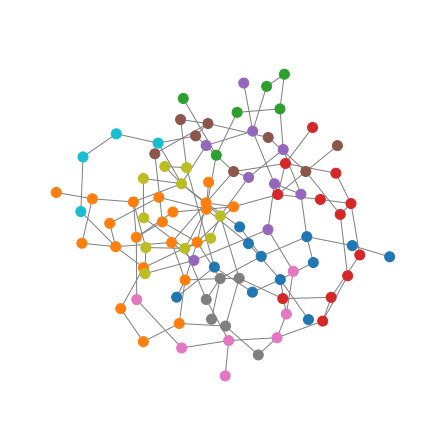}&\imgcell{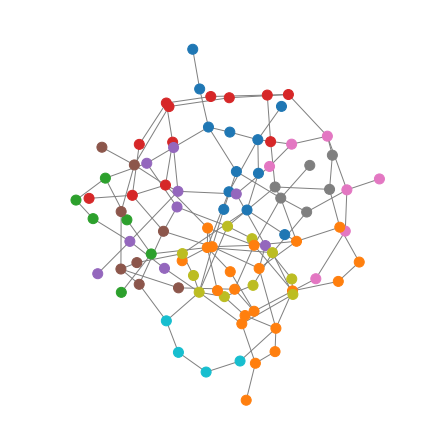}&\imgcell{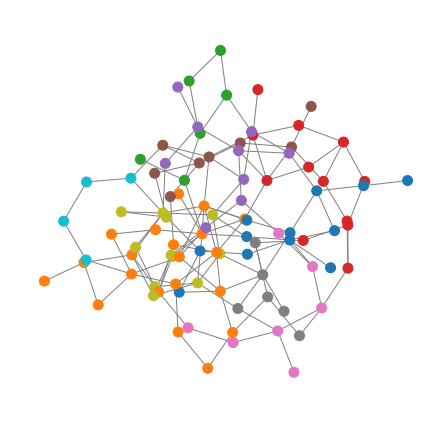}&\imgcell{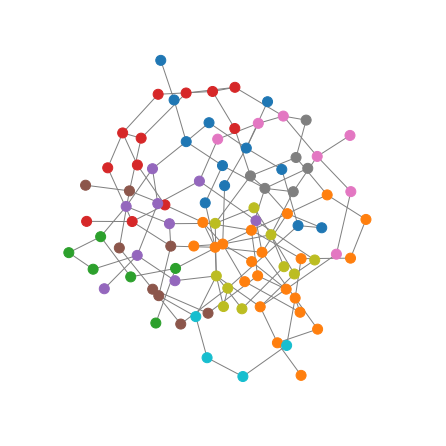}&\imgcell{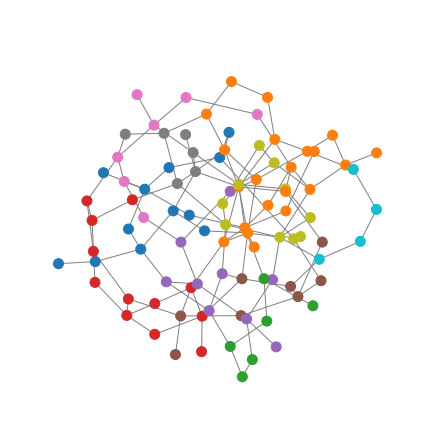}\\

\imgcell{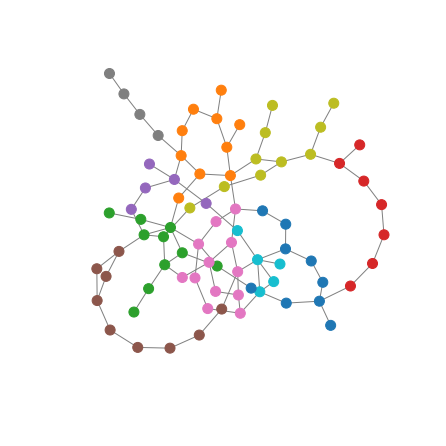}&\imgcell{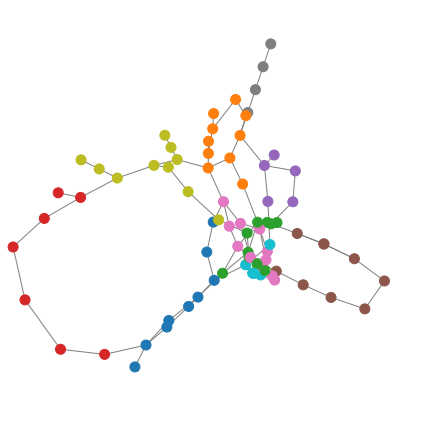} &\imgcell{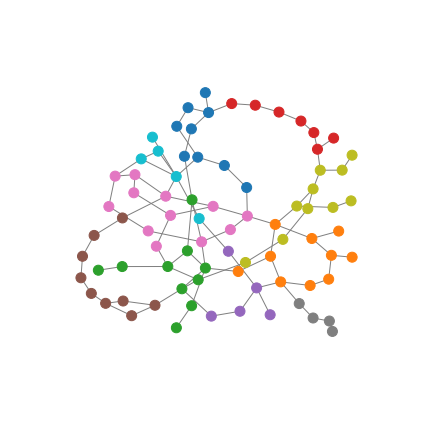}&\imgcell{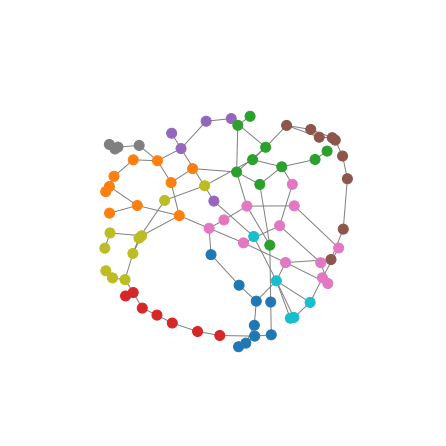}&\imgcell{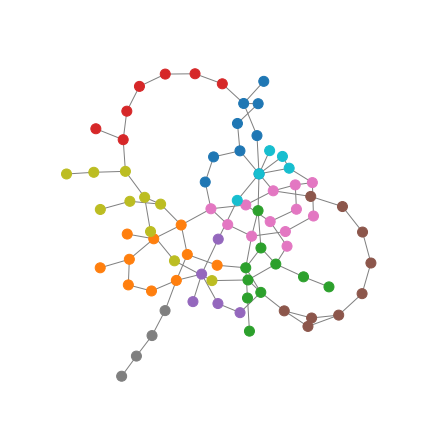}&\imgcell{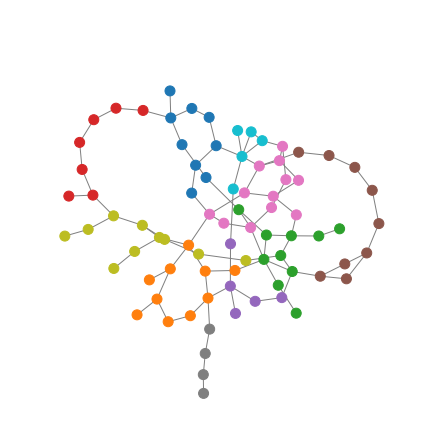}&\imgcell{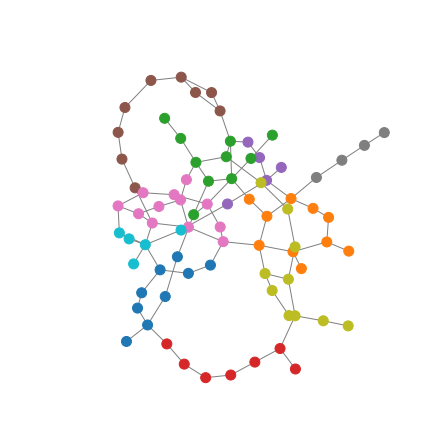}&\imgcell{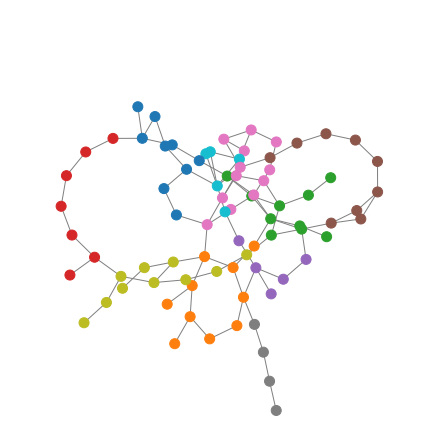}&\imgcell{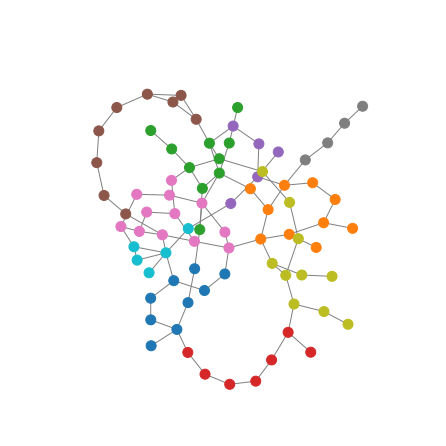}&\imgcell{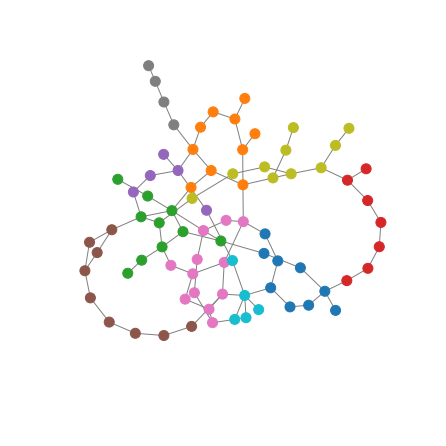}\\

\end{tabular}
\captionof{figure}[]{The qualitative evaluation of DeepGD with PivotMDS initialization. Each row corresponds to one test graph and each column represents one method. The colors of the nodes represent their community within the graph.}
\label{fig:vis-result}
\end{table*} 

\subsection{Multi-Objective Training Strategy}
\label{sec:strategy-result}
Different aesthetic metrics measure different visual properties, and optimizing one
metric sometimes cause others to worsen. For example, if we try to optimize stress and minimum angle simultaneously, the minimum angle metric will be worsened when we put more effort in optimizing stress. Hence, in this multi-objective settings, our goal is to find the Pareto optimal between two aesthetics such that no change can be made to improve both aesthetics. The question we need to answer in this section is which training strategies can help us get closer to the Pareto optimal.

\begin{figure}[htbp!]
 \centering
 \adjustbox{trim={0\width} {0\height} {0\width} {0\height},clip}{\includegraphics[width=\linewidth]{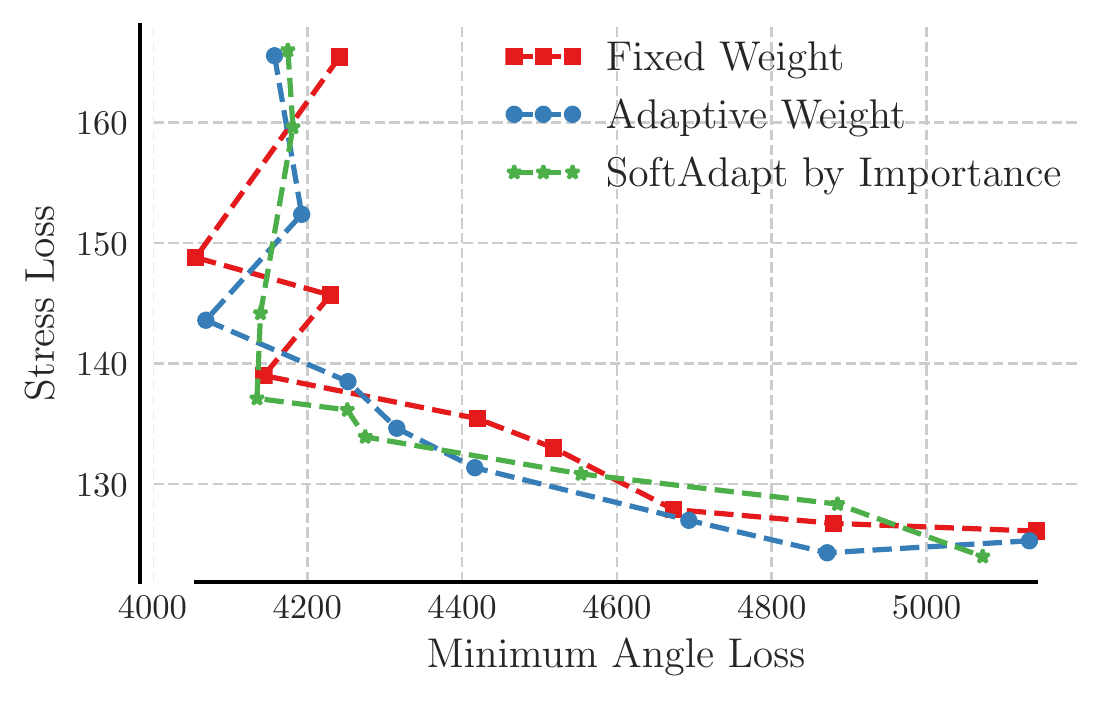}}
 \caption{The Pareto frontier for different training strategies. The points on each line represents one  model with different importance/weight factors. The connections indicate that the starting point of that connection has smaller stress weight/importance than its ending point. 
}
 \label{fig:pareto-frontier}
\end{figure}

For the two multi-objective training strategies proposed in \autoref{sec:method-strategy}, their effectiveness is assessed by comparing against the simplest training strategy of setting the fixed weight factor for each epoch. Taking stress and minimum angle as an example, the Pareto frontier lines are drawn for these three training strategies in \autoref{fig:pareto-frontier}.

As our optimization goal is to minimize both the minimum angle loss and the stress loss, the Pareto optimal should locate at the bottom left of the Pareto frontier plot. It is obvious that, the adaptive weight and SoftAdapt by importance is closer to the Pareto optimal for most of the points, in comparison with the fixed weight strategy. In conclusion, in this multi-objective optimization problem, we are more likely to achieve the Pareto optimal with adaptive weight and SoftAdapt by importance.

\subsection{Computation Time}
\label{sec:train-test-time}
For training, it takes 220 seconds on average for each epoch. According to our observations, for all models we trained, the validation loss converges after around 400 epochs. After training, the average testing time per graph is 0.049 seconds. In a word, once trained, DeepGD can generate layouts in real-time for any modest sized graphs to optimize the desired aesthetics.

% After training, the average testing time per graph is 0.049 seconds on the GPU and 0.266 seconds on the CPU. 
% However, the average testing time of Graphviz is 0.125 seconds. Therefore, we can see that DeepGD can draw a graph significantly faster than Graphviz.

\subsection{Scalability}

\xiaoqi{Even though the main focus of this paper is to draw small graphs ($\leq 100$ nodes), we still estimated the model capability on large graphs from SuiteSparse Matrix Collection\footnote{\url{https://sparse.tamu.edu/}} with thousands of nodes. Using 16GB of GPU memory, DeepGD can draw graphs with at most 4000 nodes. For 8 large graphs with 3500-4000 nodes, DeepGD can draw them in 28.14 seconds on average.

% DeepGD was tested to draw 464 large graphs with 500-4000 nodes within 39.49 minutes on GPU in total, whereas Graphviz needs 103.8 minutes on CPU. Among all large graphs, for 8 large graphs with 3500-4000 nodes, DeepGD can draw them within 3.75 minutes in total on GPU but Graphviz finish the drawing in 8.3 minutes on CPU.

In terms of the visualization quality, the performance of DeepGD cannot be guaranteed for large graphs because DeepGD is only trained on small graphs. Specifically, we observed that DeepGD performs significantly better than Graphviz for some large graphs but may fail to outperform Graphviz for some other large graphs, regardless of the graph size. 

We note that DeepGD was not trained on large graphs, due to the limitation of long training time on these graphs. Nevertheless, we believe it is possible to scale DeepGD to very large graphs by considering only the original edges in the graph plus a sparse subset of node pairs that are not neighbors. We are currently working along this direction.}

\begin{table}[hbtp!]
\centering
\fontsize{8}{8}\selectfont
\begin{tabular}{ c|cc }
    \bfseries{Graph} & \bfseries{Graphviz} & \bfseries{DeepGD} \rule[-1ex]{0pt}{0ex}\\ \hline
    \makecell{\bfseries{msc00726} \\ $n=726$} & 
    \raisebox{-0.5\totalheight}{\adjustbox{height=10em, trim={0.11\width} {0\height} {0.17\width} {0\height},clip}{\adjincludegraphics[]{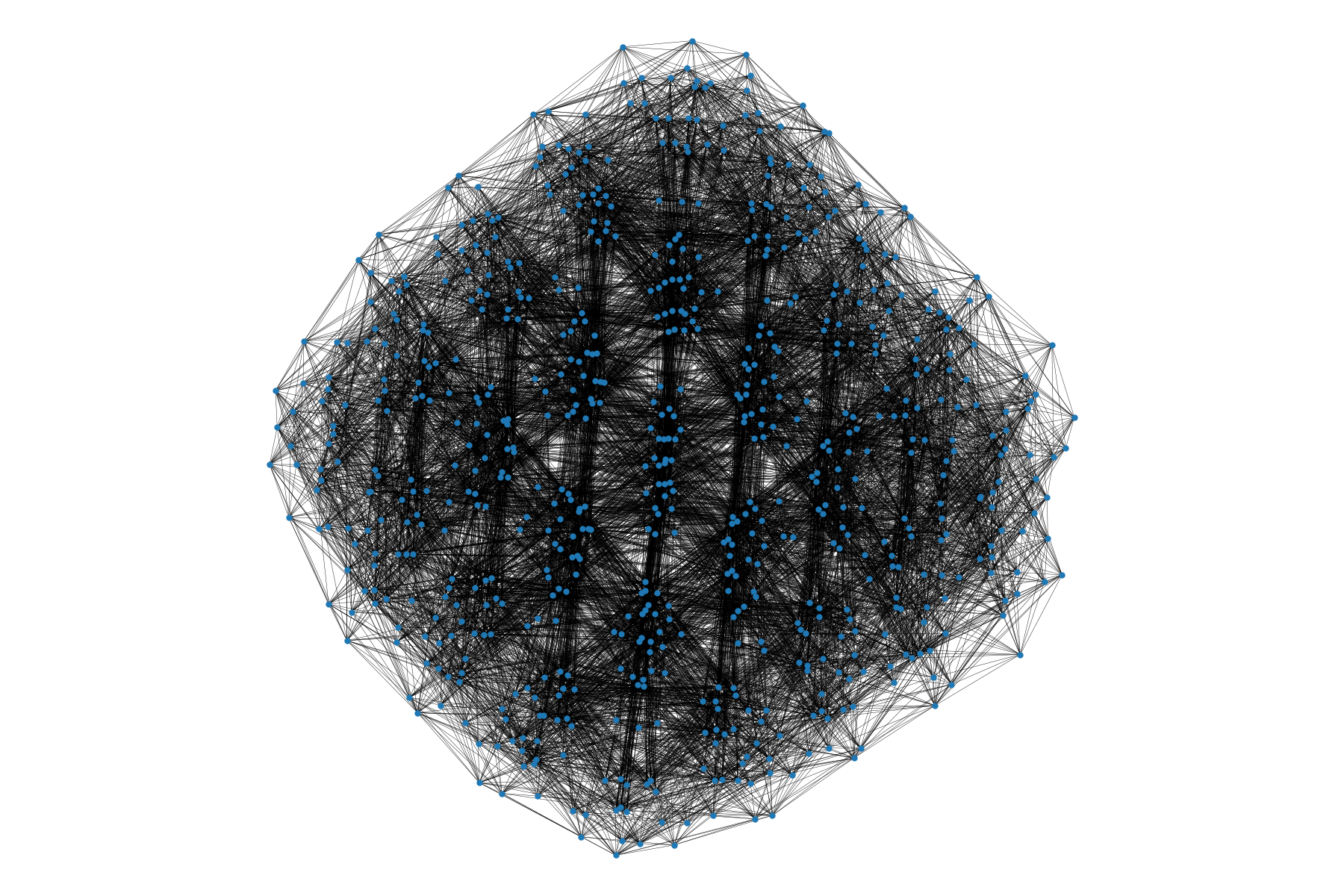}}} & 
    \raisebox{-0.5\totalheight}{\adjustbox{height=10em, trim={0.17\width} {0\height} {0.12\width} {0\height},clip}{\adjincludegraphics[]{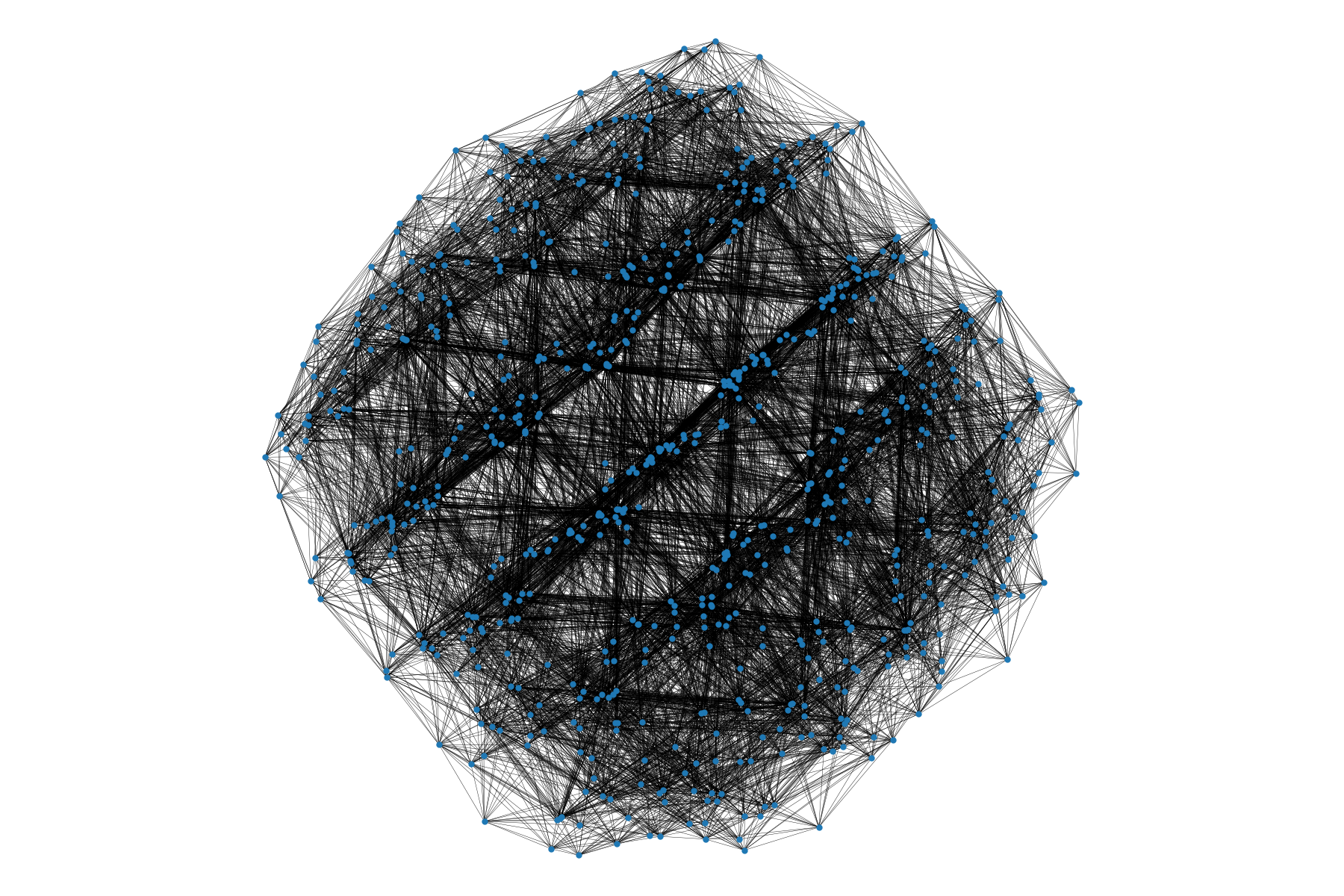}}}\\
    \makecell{\bfseries{rdist3a} \\ $n=2398$} & 
    \raisebox{-0.5\totalheight}{\adjustbox{height=10em, trim={0.11\width} {0\height} {0.17\width} {0\height},clip}{\adjincludegraphics[]{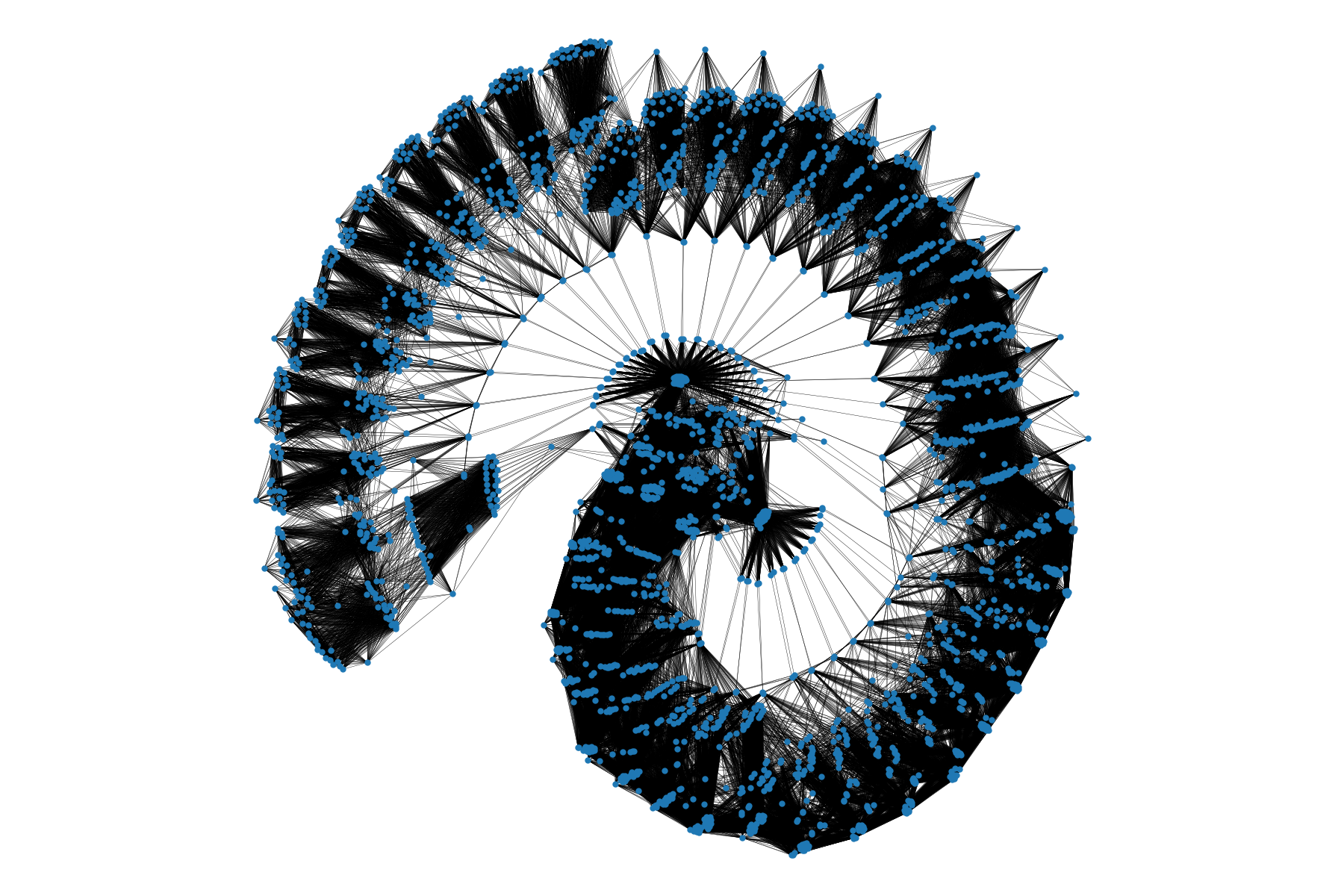}}} & 
    \raisebox{-0.5\totalheight}{\adjustbox{height=10em, trim={0.17\width} {0\height} {0.12\width} {0\height},clip}{\adjincludegraphics[]{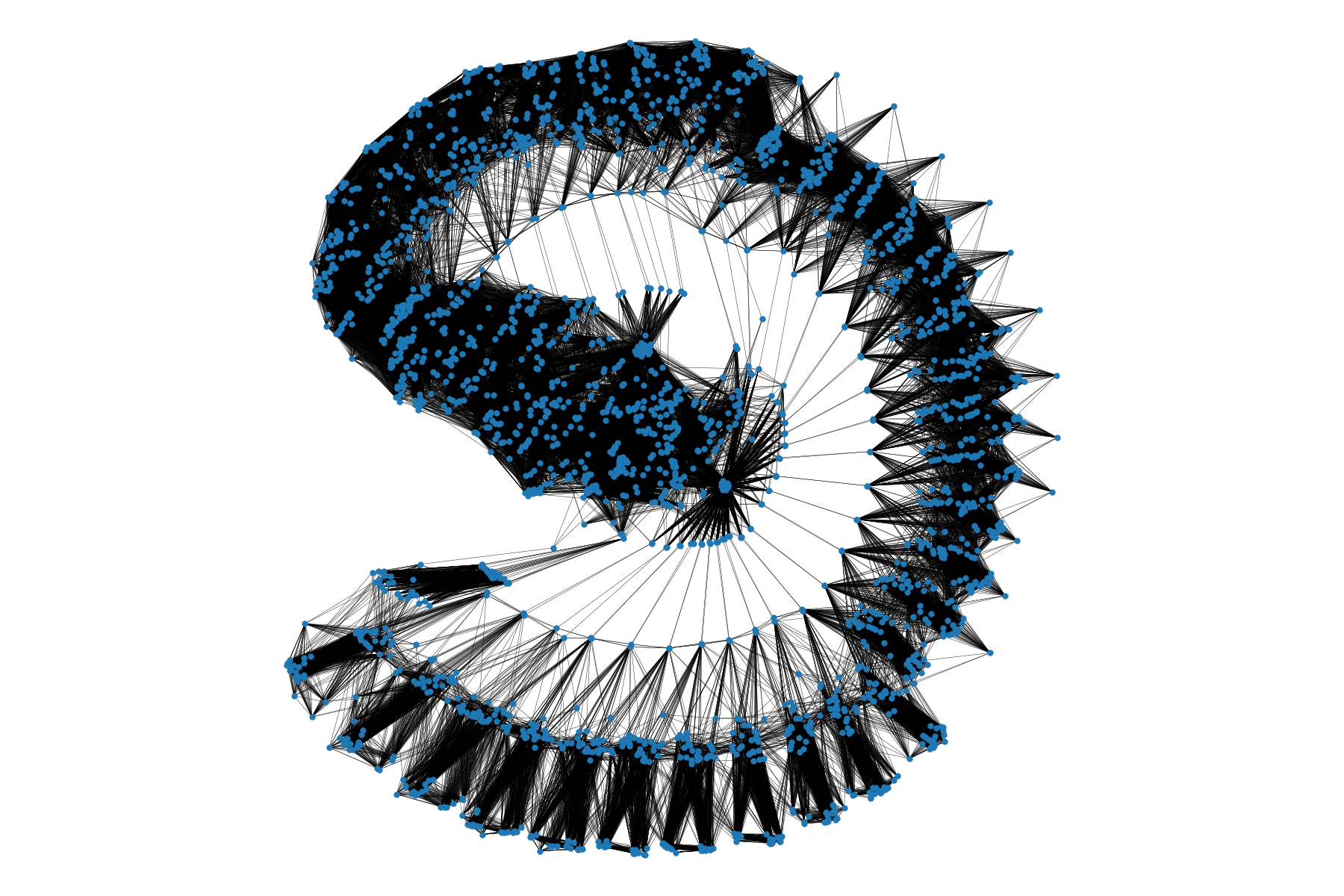}}} \\
    \makecell{\bfseries{heart1} \\ $n=3557$} & 
    \raisebox{-0.5\totalheight}{\adjustbox{height=10em, trim={0.11\width} {0\height} {0.17\width} {0\height},clip}{\adjincludegraphics[]{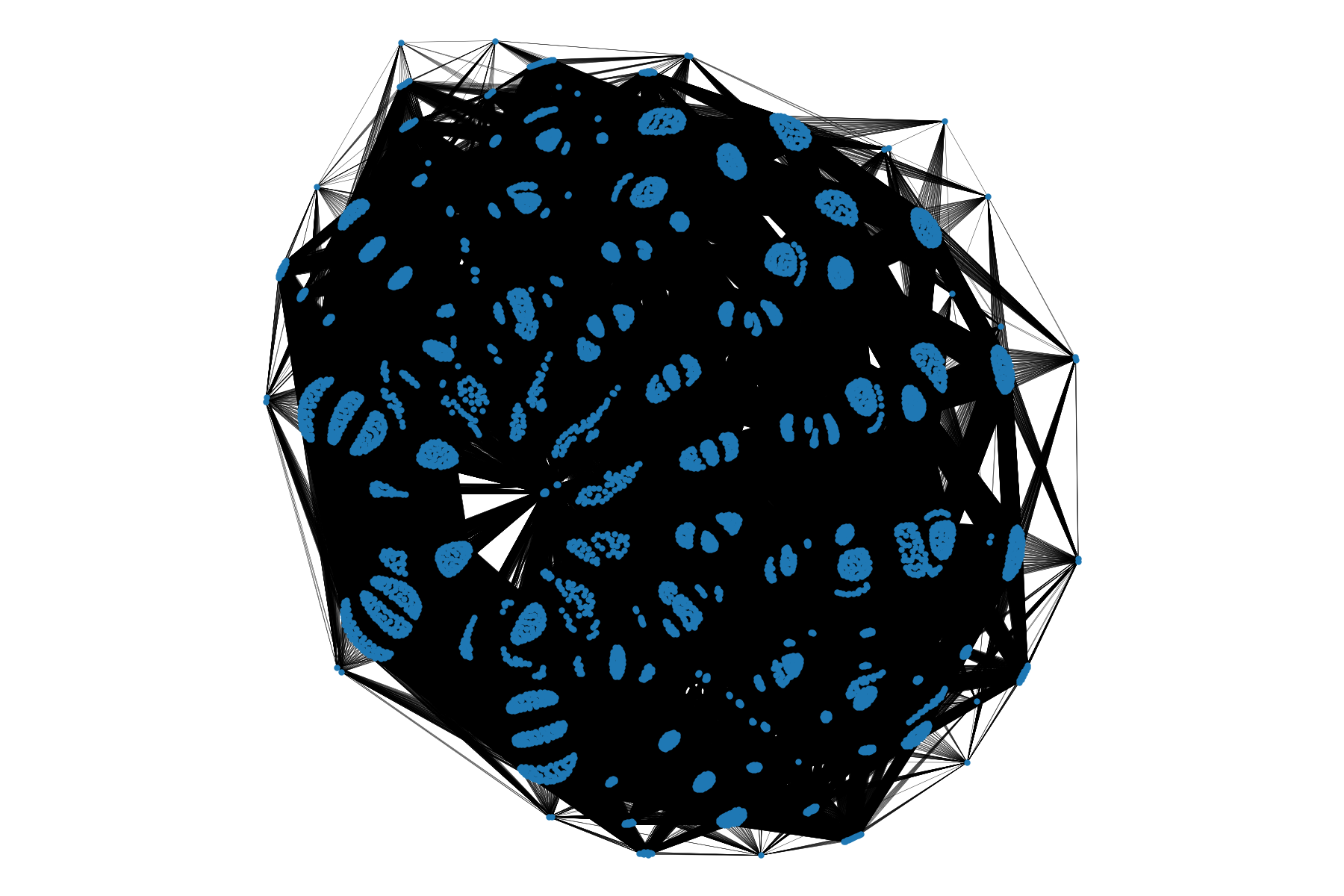}}} & 
    \raisebox{-0.5\totalheight}{\adjustbox{height=10em, trim={0.17\width} {0\height} {0.12\width} {0\height},clip}{\adjincludegraphics[]{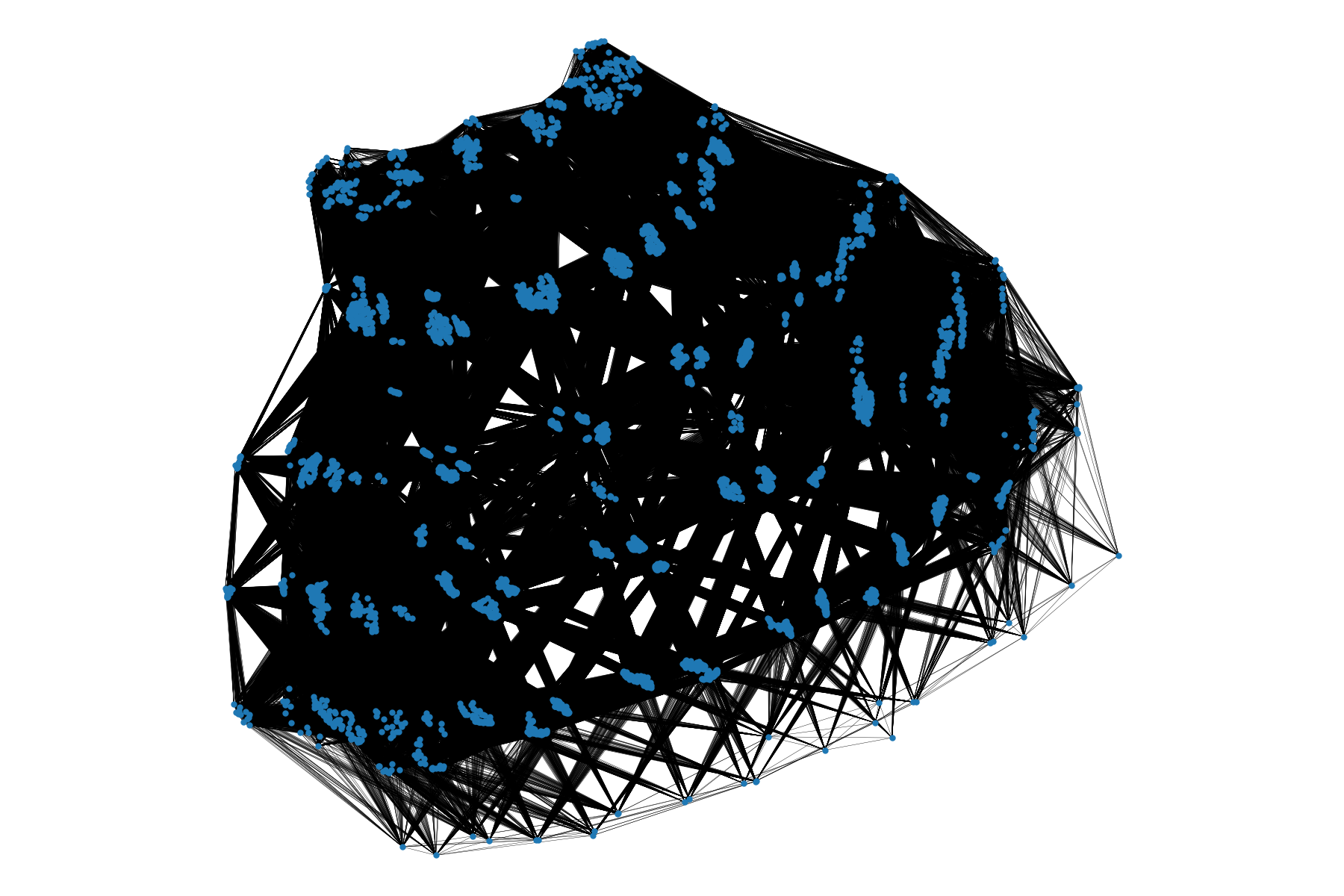}}}
\end{tabular}   
\captionof{figure}[]{The qualitative comparison between Graphviz and DeepGD over three large graphs, where $n$ denotes the number of nodes in the graph.}
\label{fig:large-result}
\end{table}

\section{Discussions}\label{sec_discussion}
The evaluation section substantiates that the DeepGD framework can generate visually pleasing layouts for unseen graphs by optimizing certain aesthetic metrics simultaneously. %Nonetheless, there are several limitations which should be addressed in the future.

% \paragraph{\textbf{Scalability}} 
%The scalability of DeepGD is mostly limited by the training data. Since Rome only contains graphs with 10 to 100 nodes, DeepGD trained with Rome might not generate reasonably good graph layouts for graphs with more than hundreds of nodes. However, because 
% It is worth to emphasize that scaling to large graph is not the main motivation of this paper. Instead, what we attempt to
% validate is whether it is possible for a deep learning based framework to produce aesthetic layout that optimize {\em arbitrary objective functions} for {\em general graphs}. A positive answer would then allow us to achieve our ultimate goal --
% employing this framework as a generator in a Generative Adversarial Network (GAN) setting, with a discriminator learned from human preference, in order to understand human perception.

%Since DeepGD requires O($N^2$) time to compute the distances for all node pairs and O($N^2$) memory to store it, the computational cost of training quadratically increases with the graph size. The fact that we are restricted to train only on small graphs also means that our model may not extend well when applied to large graphs. 
%A possible solution we plan to investigate is to only compute the distance for two nodes within a small graph theoretic distance, thus reducing the complexity of the model. %For this case, the computational complexity can be reduced but the local topological structure can still be preserved.

\paragraph{\textbf{Choice of Importance Factor}} In multi-objective settings, the importance factor is specified by the user and is supposed to reflect human preferences on different loss components. However, it's still difficult to quantify the abstract human preference into a single number, even though the difference in numerical scale is already automatically taken into account by the adaptive weight training strategy. We plan to explore models that can learn human preference automatically.
%and specify the importance factor without human input.

% \paragraph{\textbf{Initialization Effect}} During inference time, the initial node position can also affect the final layout to some extent. 
% If the node position is randomly initialized but with a fixed random seed, the generated layout is deterministic. 
% However, it is hard to find a fixed random seed which can provide us with a good starting point for all  graphs. Given the fast  inference speed of DeepGD, one solution is to do multiple runs and pick the best layouts,
% %On the contrary, if the node position is randomly initialized without a fixed random seed, the generated layout is non-deterministic and might look better with a good random initialization. So, the problem is how to find a good deterministic initialization for every test graph. One possible 
% another possibility we will explore is to start with the initial layout from a fast and deterministic graph drawing algorithm such as PivotMDS.

\paragraph{\textbf{Training Data Representativity}} A common issue of deep learning models is that the representativity of training data constrains the model capability of generalizing to unseen data. If the model does not see a specific type of graph during training, it is challenging for the model to draw that type of graph during inference. For example, we observed that DeepGD cannot draw star graph well. We suspect the reason is that Rome does not contain graphs with extremely large node degree. Therefore, we conducted an experiment of training DeepGD with Rome and North\footnote{\url{http://www.graphdrawing.org/data.html}} graph dataset together. Since North data includes many graphs with large node degree, DeepGD trained by both Rome and North can draw the star graph properly. In a word, if the training data is not representative enough, the generalization of model could be affected.
% \footnoteref{gd-data}

\section{Conclusions}
We propose a novel ConvGNN-based framework, DeepGD, which can generate graph layouts such that any desired combination of aesthetics can be complied with, as long as the aesthetics can be expressed as differentiable cost functions. To balance among multiple aesthetics, two adaptive training strategies for graph drawing are proposed to dynamically adjust the weight factors. It is worth mentioning that optimizing any aesthetic metrics only requires a redefined loss function without any algorithmic change. Compared to other deep learning based graph drawing algorithms, DeepGD only needs to be trained once and can subsequently be applied to draw arbitrary types of graphs. 

We explored the effectiveness and efficiency of DeepGD by optimizing stress, minimum angle, edge length variation, t-SNE, and node occlusion. The quantitative and qualitative evaluation demonstrate that DeepGD outperforms the baseline models on all of the five aesthetic metrics we experimented, especially for the DeepGD initialized by PivotMDS with four aesthetics. 

%\acknowledgments{
%}

%\bibliographystyle{abbrv}
\bibliographystyle{abbrv-doi}

\bibliography{reference}

\begin{thebibliography}{10}

\bibitem{PMDS}
U.~Brandes and C.~Pich.
\newblock Eigensolver methods for progressive multidimensional scaling of large
  data.
\newblock {\em LNCS}, 4372, 09 2006. doi: {{%
10\hspace{.1pt}\discretionary{.}{%
}{.}\hspace{.4pt}1007\discretionary{/}{%
}{/}978\discretionary{%
}{-}{-}3\discretionary{%
}{-}{-}540\discretionary{%
}{-}{-}70904\discretionary{%
}{-}{-}6\_6}}


\bibitem{didimo}
W.~Didimo, G.~Liotta, and S.~Romeo.
\newblock Topology-driven force-directed algorithms.
\newblock {\em Proc. of GD 2010}, 6502:165--176, 09 2010. doi: {{%
10\hspace{.1pt}\discretionary{.}{%
}{.}\hspace{.4pt}1007\discretionary{/}{%
}{/}978\discretionary{%
}{-}{-}3\discretionary{%
}{-}{-}642\discretionary{%
}{-}{-}18469\discretionary{%
}{-}{-}7\_15}}


\bibitem{gansner_stress_major}
E.~R. Gansner, Y.~Koren, and S.~North.
\newblock Graph drawing by stress majorization.
\newblock {\em Graph Drawing Lecture Notes in Computer Science}, p. 239–250,
  2005. doi: {{%
10\hspace{.1pt}\discretionary{.}{%
}{.}\hspace{.4pt}1007\discretionary{/}{%
}{/}978\discretionary{%
}{-}{-}3\discretionary{%
}{-}{-}540\discretionary{%
}{-}{-}31843\discretionary{%
}{-}{-}9\_25}}


\bibitem{gilmer-riley}
J.~Gilmer, S.~Schoenholz, P.~Riley, O.~Vinyals, and G.~Dahl.
\newblock Neural message passing for quantum chemistry.
\newblock 04 2017.

\bibitem{haleem-huamin}
H.~Haleem, Y.~Wang, A.~Puri, S.~Wadhwa, and H.~Qu.
\newblock Evaluating the readability of force directed graph layouts: A deep
  learning approach.
\newblock {\em IEEE computer graphics and applications}, 39:40--53, 07 2019.
  doi: {{%
10\hspace{.1pt}\discretionary{.}{%
}{.}\hspace{.4pt}1109\discretionary{/}{%
}{/}MCG\hspace{.1pt}\discretionary{.}{%
}{.}\hspace{.4pt}2018\hspace{.1pt}\discretionary{.}{%
}{.}\hspace{.4pt}2881501}}


\bibitem{hamilton-ying}
W.~Hamilton, R.~Ying, and J.~Leskovec.
\newblock Inductive representation learning on large graphs.
\newblock 06 2017.

\bibitem{softadapt}
A.~A. Heydari, C.~Thompson, and A.~Mehmood.
\newblock Softadapt: Techniques for adaptive loss weighting of neural networks
  with multi-part loss functions.
\newblock 12 2019.

\bibitem{huang-2013}
W.~Huang, P.~Eades, S.-H. Hong, and C.-C. Lin.
\newblock Improving multiple aesthetics produces better graph drawings.
\newblock {\em Journal of Visual Languages \& Computing}, 24:262–272, 08
  2013. doi: {{%
10\hspace{.1pt}\discretionary{.}{%
}{.}\hspace{.4pt}1016\discretionary{/}{%
}{/}j\hspace{.1pt}\discretionary{.}{%
}{.}\hspace{.4pt}jvlc\hspace{.1pt}\discretionary{.}{%
}{.}\hspace{.4pt}2011\hspace{.1pt}\discretionary{.}{%
}{.}\hspace{.4pt}12\hspace{.1pt}\discretionary{.}{%
}{.}\hspace{.4pt}002}}


\bibitem{kruiger_2017}
J.~F. Kruiger, P.~E. Rauber, R.~M. Martins, A.~Kerren, S.~Kobourov, and A.~C.
  Telea.
\newblock Graph layouts by t‐sne.
\newblock {\em Computer Graphics Forum}, 36(3):283–294, 2017. doi: {{%
10\hspace{.1pt}\discretionary{.}{%
}{.}\hspace{.4pt}1111\discretionary{/}{%
}{/}cgf\hspace{.1pt}\discretionary{.}{%
}{.}\hspace{.4pt}13187}}


\bibitem{kwon-ma-2020}
O.-H. Kwon and K.-L. Ma.
\newblock A deep generative model for graph layout.
\newblock {\em IEEE Transactions on Visualization and Computer Graphics}, 26,
  01 2020. doi: {{%
10\hspace{.1pt}\discretionary{.}{%
}{.}\hspace{.4pt}1109\discretionary{/}{%
}{/}TVCG\hspace{.1pt}\discretionary{.}{%
}{.}\hspace{.4pt}2019\hspace{.1pt}\discretionary{.}{%
}{.}\hspace{.4pt}2934396}}


\bibitem{mchedlidze}
T.~Mchedlidze, A.~Pak, and M.~Klammler.
\newblock Aesthetic discrimination of graph layouts.
\newblock {\em Journal of Graph Algorithms and Applications}, 23:525--552, 01
  2019. doi: {{%
10\hspace{.1pt}\discretionary{.}{%
}{.}\hspace{.4pt}7155\discretionary{/}{%
}{/}jgaa\hspace{.1pt}\discretionary{.}{%
}{.}\hspace{.4pt}00501}}


\bibitem{umap}
L.~McInnes, J.~Healy, N.~Saul, and L.~Grossberger.
\newblock Umap: Uniform manifold approximation and projection.
\newblock {\em Journal of Open Source Software}, 3:861, 09 2018. doi: {{%
10\hspace{.1pt}\discretionary{.}{%
}{.}\hspace{.4pt}21105\discretionary{/}{%
}{/}joss\hspace{.1pt}\discretionary{.}{%
}{.}\hspace{.4pt}00861}}


\bibitem{Nascimento}
H.~Nascimento, P.~Eades, and C.~Mendonça.
\newblock A multi-agent approach using a-teams for graph drawing.
\newblock {\em Proceedings of the 9th International Conference on Intelligent
  Systems}, pp. 39--42, 01 2000.

\bibitem{neto_eades_1993}
C.~F. X.~M. Neto and P.~Eades.
\newblock Learning aesthetics for visualization.
\newblock {\em Anais do XX Semin´ario Integrado de Software e Hardware}, p.
  76–88, 1993.

\bibitem{purchase-helen-angle}
H.~Purchase.
\newblock Metrics for graph drawing aesthetics.
\newblock {\em Journal of Visual Languages \& Computing}, 13:501--516, 10 2002.
  doi: {{%
10\hspace{.1pt}\discretionary{.}{%
}{.}\hspace{.4pt}1006\discretionary{/}{%
}{/}jvlc\hspace{.1pt}\discretionary{.}{%
}{.}\hspace{.4pt}2002\hspace{.1pt}\discretionary{.}{%
}{.}\hspace{.4pt}0232}}


\bibitem{sponemann}
M.~Spönemann, B.~Duderstadt, and R.~von Hanxleden.
\newblock Evolutionary meta layout of graphs.
\newblock 8578:16--30, 07 2014. doi: {{%
10\hspace{.1pt}\discretionary{.}{%
}{.}\hspace{.4pt}1007\discretionary{/}{%
}{/}978\discretionary{%
}{-}{-}3\discretionary{%
}{-}{-}662\discretionary{%
}{-}{-}44043\discretionary{%
}{-}{-}8\_3}}


\bibitem{VERSE}
A.~Tsitsulin, D.~Mottin, P.~Karras, and E.~Müller.
\newblock Verse: Versatile graph embeddings from similarity measures.
\newblock 03 2018. doi: {{%
10\hspace{.1pt}\discretionary{.}{%
}{.}\hspace{.4pt}1145\discretionary{/}{%
}{/}3178876\hspace{.1pt}\discretionary{.}{%
}{.}\hspace{.4pt}3186120}}


\bibitem{tsne}
L.~Van Der~Maaten and G.~Hinton.
\newblock Viualizing data using t-sne.
\newblock {\em Journal of Machine Learning Research}, 9:2579--2605, 11 2008.

\bibitem{wang-qu-2019}
Y.~Wang, Z.~Jin, Q.~Wang, W.~Cui, T.~Ma, and H.~Qu.
\newblock Deepdrawing: A deep learning approach to graph drawing.
\newblock {\em IEEE Transactions on Visualization and Computer Graphics},
  PP:1--1, 08 2019. doi: {{%
10\hspace{.1pt}\discretionary{.}{%
}{.}\hspace{.4pt}1109\discretionary{/}{%
}{/}TVCG\hspace{.1pt}\discretionary{.}{%
}{.}\hspace{.4pt}2019\hspace{.1pt}\discretionary{.}{%
}{.}\hspace{.4pt}2934798}}


\end{thebibliography}

\newpage
\appendixpage
\label{appendices}
\appendix
\begin{appendices}

\section{Model Configuration}
\label{sec:model-config}
%In terms of model configuration, o
Our model was trained in mini-batches of 128 graphs. We used AdamW optimizer with a weight decay rate of 0.01 in order to shrink the model parameters for each optimization step as a form of regularization. Regarding the learning rate, it was initially set to 0.01 and then decayed exponentially after each epoch with a rate of 0.99. In this case, smaller steps were taken as the model approached the local minimum of the loss function. In addition, batch normalization was applied for each hidden layer with the purpose of making optimization landscape significantly smoother. For activation function, LeakyReLU was used for each hidden layer in order to alleviate the vanishing gradient problem.

Our model architecture consists of 9 residual blocks and each layer in the blocks has 8 neurons. For each convolutional layer, there is a edge feature network with 2 hidden layer to process information from edge feature. In total, the model has 30 hidden layers and 42290 model parameters. The model takes a complete digraph (including virtual edges) as input. The experimental results for comparing different model architectures are presented in \autoref{sec:model-archi-result}.

\section{Model Architectures Comparison}
\label{sec:model-archi-result}
We conducted a series of experiments to compare different model architectures, which include:
\begin{itemize}
\item Understood the effect of different numbers of neuron for each layer in the residual blocks.
\item Observed the impact of different number of hidden layers in the edge feature network.
\item Conducted ablation study of edge features for residual blocks.
\item Explored the effect of removing residual connection.
\item Discussed the impact of using original graph instead of complete graph.

\end{itemize}
All the experiments presented below were conducted on DeepGD with stress loss only and PMDS initialization. 

\subsection{Edge Feature Networks} We first compare the effect of having different number of hidden layers in the edge feature network. The number of hidden layers in the edge feature network is the key factor for the computational complexity of DeepGD because a separate edge feature network is trained for each ConvGNN layer. Hence, our goal is to find the minimum number of hidden layers in the edge feature network such that the model performance is not compromised too much. We can clearly see that the edge feature network with 2 layers can find a good balance between performance and the number of parameters because it only 0.23\% worse than edge feature network with 3 layers. Therefore, considering the computational cost, the edge feature network in our 
final model architecture contains 2 hidden layers.

\begin{table}[ht]
\renewcommand{\arraystretch}{1.4}
\setlength{\tabcolsep}{7pt}
\caption{Comparison of different number of hidden layers in the edge feature network. }
\label{table:layer-edgenet}
\centering
\renewcommand\theadfont{\normalsize\bfseries}
\begin{tabular}{c|c|c|c}
\textbf{Number of Hidden Layers} & 1 & 2 & 3\\ 
\hline
\textbf{Stress} & 243.45 & \textbf{239.73} & 239.33\\  \hline  
\textbf{Stress SPC w.r.t. Graphviz} & -2.19\% & \textbf{-5.43\%} & -5.66\%\\    
\end{tabular}
\end{table}

\subsection{Number of Hidden Neuron for Residual Blocks} Given that each residual block has 3 hidden layers, the number of neurons for those three layers is a key component for model architecture design. Therefore, we carefully explored the effect of different numbers of hidden neurons. The experiments we conducted include (8,8,8), (16,16,8), and (8,8,2), where the $i$th number in the tuple denotes the number of neurons in the $i$th layer. The three DeepGD models presented in \autoref{table:hidden-neuron} incorporate normalized direction and Euclidean distance as the two extra edge features for the residual blocks, and have one hidden layer in the edge feature network. The result shows that the (8,8,8) model with less number of parameters can achieve comparably good performance to (16,16,8). Hence, for our best model architecture, we adopt (8,8,8) as the number of hidden neurons in all residual blocks.

\begin{table}[ht]
\renewcommand{\arraystretch}{1.4}
\setlength{\tabcolsep}{6pt}
\caption{Comparison of different numbers of neurons in the residual blocks. }
\label{table:hidden-neuron}
\centering
\renewcommand\theadfont{\bfseries}
\begin{tabular}{c|c|c|c}
\textbf{Number of Hidden Neuron} & (8,8,8) & (16,16,8) & (8,8,2)\\ 
\hline
\textbf{Stress} & \textbf{239.73} & 239.49 & 242.65 \\  \hline  
\textbf{Stress SPC w.r.t. Graphviz} & \textbf{-5.43\%} & -5.44\% & -3.00\% \\    
\end{tabular}
\end{table}

\subsection{Edge Features for Residual Blocks} As stated in the Methodology section of our manuscript, two additional edge features are added as the input for each residual block.  Inspired by the stress majorization approach, we  conducted a ablation study to quantitatively assess the effectiveness of removing those two additional features (see \autoref{table:residual-blocks}). In terms of the experimental settings for the four models in \autoref{table:residual-blocks}, there are 8 neurons for each hidden layer in the residual blocks and the edge feature network contains two layer. Obviously, the DeepGD model with both Euclidean distance and normalized direction as two additional edge features significantly outperforms the other three DeepGD models. 
\begin{center}
\begin{table}[ht]
\renewcommand{\arraystretch}{1.4}
\setlength{\tabcolsep}{3.8pt}
\caption{Ablation study of edge features to the residual blocks.}
\label{table:residual-blocks}

\renewcommand\theadfont{\bfseries}

\begin{tabular}{c|c|c}
\thead{Edge Features for Residual Blocks} & \thead{Stress} & \thead{Stress SPC\\ w.r.t. Graphviz}\\ 
                    \hline
 No Additional Features & 260.96 & 10.59\%\\ 
 \hline
Direction Only & 244.63 & -0.83\% \\    
\hline
Euclidean Distance Only& 247.62 & 2.33\% \\    
\hline
Euclidean Distance and Direction & \textbf{239.73} & \textbf{-5.43\%} \\    
\end{tabular}

\end{table}
\end{center}

\end{appendices}

\end{document}